\documentclass[10pt,twocolumn,letterpaper]{article}

\usepackage[pagenumbers]{cvpr} % To force page numbers, e.g. for an arXiv version
\definecolor{bottlegreen}{HTML}{006A4E}

\usepackage{tocloft}
\usepackage{booktabs}
\usepackage{multirow}
\usepackage{graphicx}
\usepackage{array}
\usepackage[most]{tcolorbox}
\usepackage[table]{xcolor}
\usepackage{pifont}
\usepackage{pmboxdraw}
\usepackage{fontawesome5}
\usepackage{tabularx}
\usepackage{wasysym}
\usepackage{array}

\newcommand{\cmark}{\textcolor{ForestGreen}{\ding{51}}}
\newcommand{\xmark}{\textcolor{BrickRed!55}{\ding{55}}}

\newcolumntype{Y}{>{\centering\arraybackslash}X}

\definecolor{cvprblue}{rgb}{0.21,0.49,0.74}
\usepackage[pagebackref,breaklinks,colorlinks,allcolors=cvprblue]{hyperref}

\def\confName{CVPR}
\def\confYear{2027}
\def\ourdata{GenAV-1.5M}
\def\ourtest{GenAV-Test}
\def\ourtestLarge{GenAV-Test-Large}

\title{
{\ourdata}\,%
\raisebox{-0.2\height}{\includegraphics[height=1.2em]{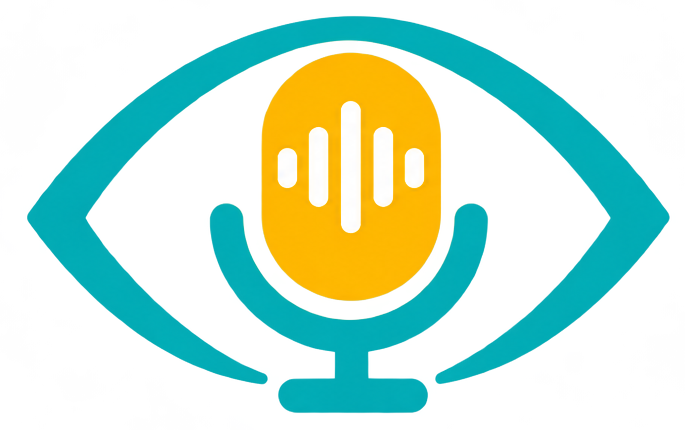}}:
Scaling In-the-Wild Audio-Visual Data for Generation
}
\author{
\textbf{Saksham Singh Kushwaha}$^{1*}$ \quad
\textbf{Preet Mukeshkumar Sojitra}$^{1*}$ \quad
\textbf{Ahnaf Adib}$^{1}$ \quad
\textbf{Weiguo Pian}$^{2,3}$ \\
\textbf{Harsh Singh}$^{4}$ \quad
\textbf{Zijun Cui}$^{1}$ \quad
\textbf{Jia Li}$^{1}$ \quad
\textbf{Sarthak Kumar Maharana}$^{1}$ \quad
\textbf{Lujing Xie}$^{1}$ \\
\textbf{Shijian Deng}$^{1}$ \quad
\textbf{Kai Wang}$^{5}$ \quad
\textbf{Latifur Khan}$^{1}$ \quad
\textbf{Wei Xiong}$^{6}$ \quad
\textbf{Yapeng Tian}$^{1}$ \\[4pt]
$^{1}$The University of Texas at Dallas \quad
$^{2}$ELLIS Institute Finland \quad
$^{3}$Aalto University \\
$^{4}$Stony Brook University \quad
$^{5}$University of Toronto \quad
$^{6}$NVIDIA \\[4pt]
$^{*}$Equal contribution \\ [4pt]
Project page: 
\href{https://sakshamsingh1.github.io/genav_webpage/}
{\textcolor{blue!50!black}{\texttt{https://sakshamsingh1.github.io/genav\_webpage/}}}
}

\begin{document}
\maketitle

\addtocontents{toc}{\protect\setcounter{tocdepth}{-1}}

\begin{abstract}

Recent advances in audio-visual generation have demonstrated impressive capabilities, yet large-scale, high-quality training data with reliable audio-visual correspondence remains scarce and largely inaccessible. We introduce \textbf{\ourdata{}}, an in-the-wild dataset of 1.54 million audio-visual clips spanning 182 sound-source categories, with rich joint captions. Curating such data at scale is challenging, as audible events may lack visible sources, while speech and music can obscure meaningful audio-visual correspondence. To address these challenges, we develop a scalable, cost-effective pipeline combining zero-shot unimodal filtering, selective audio suppression, and joint audio-visual correspondence verification using open-source models. We evaluate \ourdata{} across video-to-audio, audio-to-video, and joint audio-visual generation. Models trained on our dataset achieve substantial improvements in generation quality, reducing Fr\'echet Audio Distance from 13.85 to 7.88 for video-to-audio generation and Fr\'echet Video Distance from 251.08 to 50.11 and 114.10 to 70.52 for audio-to-video and joint generation, respectively, relative to established baselines on \ourtest{}. Additional evaluations demonstrate strong performance on external benchmarks, while scaling studies and human assessments further validate the dataset's usefulness. We will release \ourdata{} and our curation pipeline to advance research in audio-visual generation.
\end{abstract}
    
\section{Introduction}
\begin{figure}[t]
    \centerline{\includegraphics[width=\linewidth]{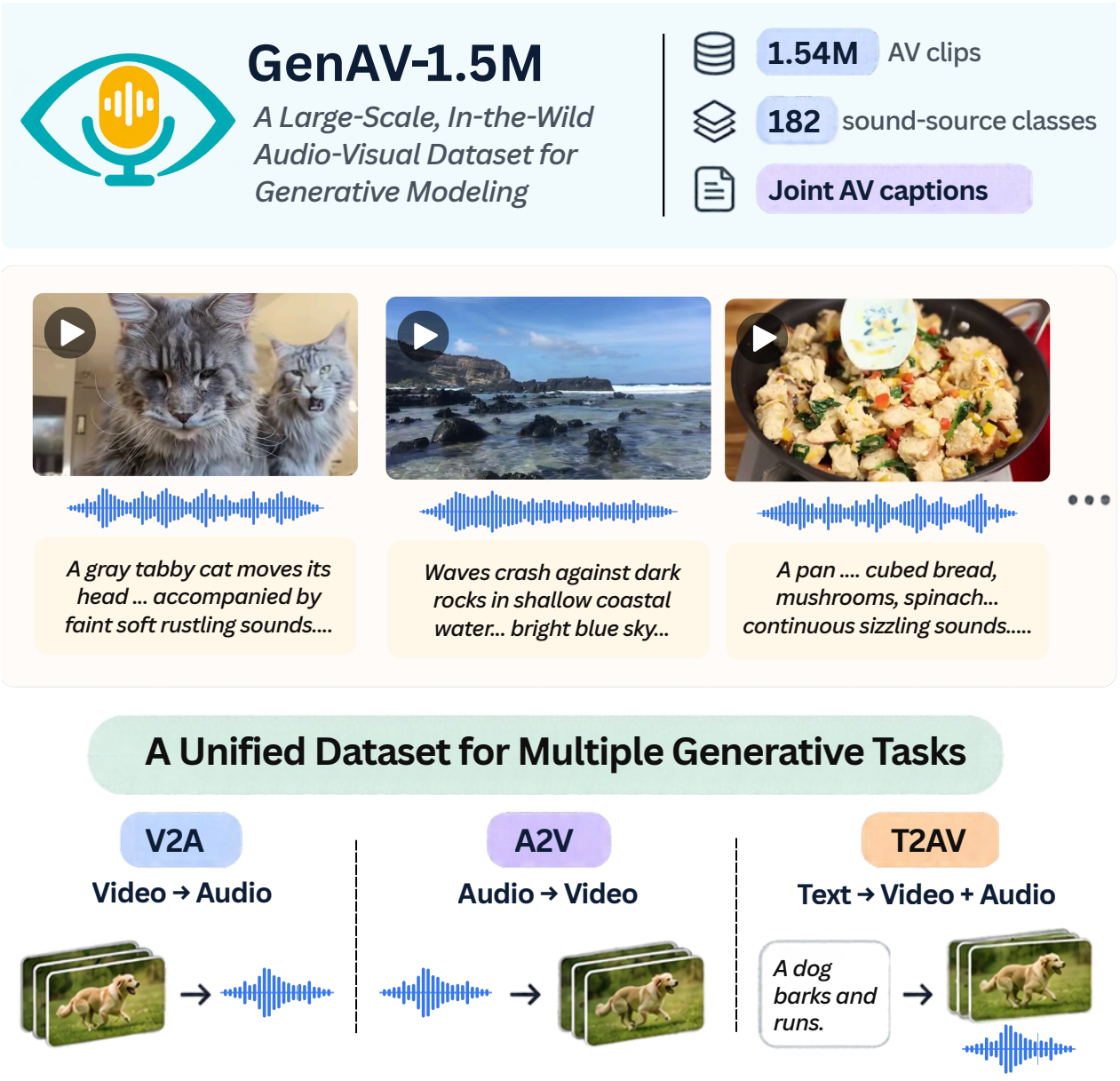}}
    % \vspace{-4.8mm}
    % \vspace{-3.0mm}
    \caption{\textbf{Overview of our \ourdata.}
    We release a high-quality, in-the-wild dataset of 1.54M audio-visual clips with rich and descriptive captions, supporting video-to-audio (\textit{V2A}), audio-to-video (\textit{A2V}), and joint audio-visual (\textit{JAV}) generation.}
    \label{fig:teaser}
    % \vspace{-5mm}
\end{figure}
%
%%%%%%%%%%%%%%%%%%%%%%
%
\begin{table*}[t]
\centering
\small
% \vspace{-5pt}
\setlength{\tabcolsep}{3pt}
\resizebox{0.8\textwidth}{!}{
\renewcommand{\arraystretch}{1.0}
\rowcolors{3}{white}{gray!8}
\begin{tabular}{@{}l@{\hspace{8pt}}cc@{\hspace{10pt}}cc@{\hspace{10pt}}cc@{\hspace{10pt}}c@{\hspace{10pt}}ccc@{}}
\toprule
& \multicolumn{2}{c}{\normalsize \textcolor{Magenta}{\textit{\textbf{Scale}}}}
& \multicolumn{2}{c}{\normalsize \textcolor{Magenta}{\textit{\textbf{Coverage}}}}
& \multicolumn{2}{c}{\normalsize \textcolor{Magenta}{\textit{\textbf{Text}}}}
& \multicolumn{1}{c}{}
& \multicolumn{3}{c}{\normalsize \textcolor{Magenta}{\textit{\textbf{Task Support}}}} \\
\cmidrule(lr){2-3} \cmidrule(lr){4-5} \cmidrule(lr){6-7} \cmidrule(lr){9-11}
\rowcolor{NavyBlue!10}
\textbf{Dataset}
& \textbf{\#Clips}
& \textbf{\#Class}
& \textbf{Open-domain}
& \textbf{AV-Corr.}
& \textbf{Captions}
& \textbf{Transcripts}
& \textbf{Annotation}
& \textbf{V2A}
& \textbf{A2V}
& \textbf{JAV} \\
\midrule

VGGSound~\cite{chen2020vggsound}
& 200K & 309 & \cmark & \cmark
& \xmark & \xmark & \faRobot\,\faUser
& \cmark & \cmark & \cmark \\

\textSFviii\textSFx~AVSync15~\cite{zhang2024audio0synchronized}
& 1,500 & 15 & \xmark & \cmark
& \xmark & \xmark & \faRobot\,\faUser
& \cmark & \cmark & \cmark \\

\textSFviii\textSFx~AVSync48~\cite{zhang2025scaling}
& 1,440 & 48 & \xmark & \cmark
& \xmark & \xmark & \faUser
& \xmark & \cmark & \xmark \\

\textSFii\textSFx~SoundAtlas-VGG~\cite{Dai_2026_CVPR}
& 140K & -- & \cmark & \cmark
& \cmark & \xmark & \faRobot
& \cmark & \xmark & \xmark \\

AudioSet~\cite{Gemmeke2017}
& 2.08M & 527 & \cmark & \xmark
& \xmark & \xmark & \faUser
& \cmark & \cmark & \cmark \\

\textSFviii\textSFx~TAVGBench~\cite{mao2024tavgbench}
& 1.7M & -- & \cmark & \xmark
& \cmark & \xmark & \faRobot
& \xmark & \xmark & \cmark \\

\textSFviii\textSFx~SoundAtlas-AS~\cite{Dai_2026_CVPR}
& 330K & -- & \cmark & \cmark
& \cmark & \xmark & \faRobot
& \cmark & \xmark & \xmark \\

\textSFviii\textSFx~VAS~\cite{chen2020generating}
& 13K & 8 & \xmark & \cmark
& \xmark & \xmark & \faUser
& \cmark & \xmark & \xmark \\

\textSFii\textSFx~VEGAS~\cite{zhou2017visual}
& 28,109 & 10 & \xmark & \cmark
& \xmark & \xmark & \faUser
& \cmark & \xmark & \xmark \\

LRS3~\cite{afouras2018lrs3}
& 151,819 & -- & \xmark & \cmark
& \xmark & \cmark & \faRobot\,\faUser
& \cmark & \cmark & \xmark \\

CommonVoice~\cite{ardila2020common}
& -- & -- & \xmark & \xmark
& \xmark & \cmark & \faUser
& \xmark & \xmark & \xmark \\

Landscape~\cite{lee2022sound0guided}
& 9,280 & 9 & \xmark & \xmark
& \xmark & \xmark & --
& \cmark & \cmark & \cmark \\

AIST++~\cite{li2021learn}
& 1,408 & 10 & \xmark & \xmark
& \xmark & \xmark & \faRobot\,\faUser
& \cmark & \cmark & \cmark \\

FAVDBench~\cite{shen2023fine0grained}
& 11,424 & 71 & \xmark & \cmark
& \cmark & \xmark & \faUser
& \xmark & \xmark & \xmark \\

MMTrail-2M~\cite{chi2024mmtrail0}
& 2M & -- & \xmark & \xmark
& \cmark & \cmark & \faRobot
& \cmark & \xmark & \cmark \\

GreatestHits~\cite{owens2016visually}
& 977 & -- & \xmark & \cmark
& \xmark & \xmark & \faUser
& \cmark & \cmark & \cmark \\

\midrule
\rowcolor{bottlegreen!13}
\textbf{\ourdata~(Ours)}
& \textbf{1.54M} & \textbf{182} & \cmark & \cmark
& \cmark & \cmark & \faRobot\,\faUser
& \cmark & \cmark & \cmark \\

\bottomrule
\end{tabular}}
\caption{
\textbf{Comparison of our proposed {\ourdata} vs. existing audio-visual generation datasets.} We release temporally-coherent 1.54M audio-visual clips along with rich captions and transcriptions, suitable for generation tasks.  AV-Corr. indicates an explicit audio-visual correspondence or consistency verification step
during dataset construction. \cmark/\xmark~indicate reported/not reported support. \faUser~and \faRobot~denote human and automated annotation, respectively. }
\label{tab:data_comparison}
% \vspace{-6mm}
\end{table*}
%
%%%%%%%%%%%%%%%%%%%%%%%%%%
Recent advances in generative modeling have driven progress from unimodal image~\cite{rombach2022high,ramesh2022hierarchical,podell2024sdxl}, video~\cite{ho2022video,hong2022cogvideo,singer2022make}, and audio~\cite{kong2020diffwave,huang2023make,liu2023audioldm} synthesis toward unified multimodal generation~\cite{liu2025javisdit,suzuki2022survey}. In particular, cross-modal and joint audio-visual generation~\cite{wang2024av,ruan2022mmdiffusion,xing2024seeing,liu2026javisdit++} requires models to capture not only visual dynamics and acoustic characteristics, but also the semantic and temporal correspondence between visible events and their sounds~\cite{cheng2025taming}. Recent frontier models~\cite{bytedance2026seedance25,minimax2026h3,deepmind2025veo31web} have demonstrated impressive audio-visual generation capabilities, yet their large-scale training datasets remain largely inaccessible. Consequently, developing open audio-visual generation models remains challenging because large-scale, high-quality training data that capture diverse real-world sound-producing events and their cross-modal correspondence are scarce.

Existing audio-visual datasets face several limitations for open-domain generation. Foundational datasets such as AudioSet~\cite{Gemmeke2017} and VGGSound~\cite{chen2020vggsound} were primarily developed for recognition, providing categorical labels rather than rich descriptions of audio-visual events. They may also contain weakly associated audio-visual pairs~\cite{morgado2021robust}, while other datasets focus on restricted domains or sound categories with limited synchronized object dynamics~\cite{lee2022sound0guided,tsuchida2019aist}. Stricter filtering for audio-visual synchronization can improve correspondence at the expense of scale and diversity; for example, AVSync15~\cite{zhang2024audio0synchronized}, curated from VGGSound, contains only 1,350 training clips across 15 categories. Recent generation-oriented efforts such as TAVGBench~\cite{mao2024tavgbench} and SoundAtlas~\cite{Dai_2026_CVPR} provide richer multimodal annotations but largely build on existing datasets, inheriting their coverage limitations. Larger multimodal datasets~\cite{chen2023valor,chen2023vast,lee2021acav100m,hochlehnert2026laion,maharana2025avrobustbench} have advanced audio-visual understanding and representation learning, but primarily target multimodal pretraining or global cross-modal alignment rather than explicitly verifying correspondence between audible events and visible sound sources for generative modeling.

Curating high-quality audio-visual data from unconstrained videos is challenging because meaningful correspondences are often sparse. Visible sound sources may be silent, audible events may originate off-screen, and speech, music, or edited soundtracks may obscure genuine audio-visual relationships. Moreover, independently detecting audible events and visible sources does not establish their correspondence, while applying expensive multimodal verification to millions of clips is computationally costly. These challenges underscore the pressing need for \textit{openly accessible, carefully curated audio-visual datasets that support diverse and temporally coherent generation}.

Fortunately, advances in pretrained multimodal models have improved zero-shot audio, visual, and cross-modal understanding~\cite{vyas2026pushing,xu2025qwen3}, enabling scalable data curation. We leverage these capabilities to develop a cost-effective, multi-stage pipeline using open-source models and targeted human verification. A flexible sound-source taxonomy guides the retrieval of diverse sound-producing events without enforcing mutually exclusive labels. Efficient zero-shot audio-text and video-text filtering first identifies relevant sounds and visible sources, followed by joint audio-visual verification to establish their correspondence. We selectively suppress interfering speech and music during verification to preserve potentially useful clips and generate rich joint captions describing visible actions, audible events, and their relationships.

\begin{figure*}[h!]\centerline{\includegraphics[width=0.9\linewidth]{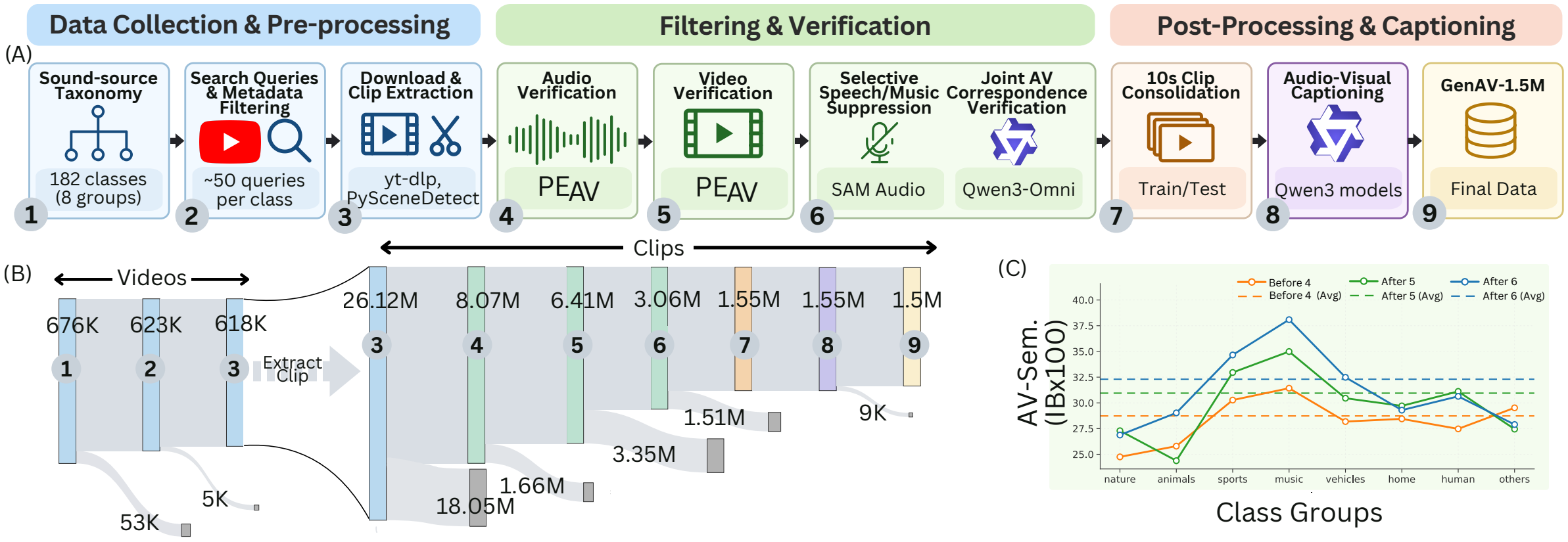}}
% \vspace{-5pt}
% \vspace{-2mm}
    \caption{\textbf{Data curation pipeline of {\ourdata}.}
    (A) Overview of data collection, three-stage audio-visual verification with selective speech/music suppression, clip consolidation, and captioning.
    (B) Number of videos/clips retained and rejected at each stage.
    (C) Audio-visual semantic similarity (IB is Imagebind~\cite{girdhar2023imagebind}) across class groups, showing improved average correspondence after filtering.}
    \label{fig:pipeline}
    % \vspace{-3mm}
\end{figure*}

Using this pipeline, we introduce \textbf{\ourdata}, comprising 1.54 million in-the-wild audio-visual clips spanning 182 sound-source categories. As illustrated in Fig.~\ref{fig:teaser}, each clip is accompanied by a rich joint audio-visual caption. We evaluate the dataset across video-to-audio (\textit{V2A}), audio-to-video (\textit{A2V}), and joint audio-visual (\textit{JAV}) generation. Models trained on \ourdata{} achieve substantial improvements in generation quality across all three tasks, with strong performance on external benchmarks. Controlled training-data comparisons, data and model scaling studies, and human caption evaluations further demonstrate the effectiveness of \ourdata{} as a training resource for audio-visual generation. Our main contributions are:

\noindent $\bullet$ \textbf{Scalable Curation Pipeline.}
We develop a scalable, cost-effective pipeline combining efficient unimodal filtering, audio-visual correspondence verification, selective audio suppression, and multimodal captioning to curate high-quality in-the-wild audio-visual data.

\noindent $\bullet$ \textbf{\ourdata~Dataset.}
We introduce a large-scale dataset of 1.54M audio-visual clips spanning 182 sound-source categories, paired with rich joint captions describing visible actions, audible events, and their relationships.

\noindent $\bullet$ \textbf{Comprehensive Evaluation.}
We demonstrate the usefulness of \ourdata~across three generation tasks through comparisons with existing models, external benchmark evaluations, controlled training-data comparisons, and data and model scaling studies.

\section{Related Works}

\noindent \textbf{Audio-Visual Generation}.
Audio-visual generation~\cite{qin2026audio} aims to synthesize audio, video, or both from multimodal inputs while maintaining semantic and temporal coherence. Existing approaches broadly fall into \textit{cross-modal} and \textit{joint audio-visual generation}. Cross-modal approaches synthesize one modality conditioned on another. Early video-to-audio (\textit{V2A}) methods focused on visually guided sound synthesis~\cite{zhang2024foleycrafter,wang2024frieren,iashin2021taming,luo2023diff}, while recent approaches leverage multimodal representations and joint audio-text/video training to improve semantic alignment and synchronization~\cite{cheng2025taming,shan2025hunyuanvideofoleymultimodaldiffusionrepresentation,Dai_2026_CVPR,kushwaha2025vintage,pian2026omnisonic}. Conversely, audio-to-video (\textit{A2V}) methods exploit acoustic semantics and temporal cues to guide video generation~\cite{jeong2023power,yariv2023diverse,linz2024asva,kim2025semantically}. Joint audio-visual generation (\textit{JAV}) models combine both modalities to synthesize coherent, synchronized audio-video content~\cite{ruan2022mmdiffusion,low2025ovi,wang2025universe,liu2026javisdit++}. Despite these advances, many generative models still rely on recognition-oriented datasets~\cite{chen2020vggsound,Gemmeke2017} adapted through filtering~\cite{zhang2024audio0synchronized} and captioning~\cite{Dai_2026_CVPR}. We evaluate \ourdata{} across all three tasks to assess its effectiveness as a general-purpose training resource for audio-visual generation.

\noindent \textbf{Audio-Visual Generation Datasets}.
Table~\ref{tab:data_comparison} compares existing datasets used for audio-visual generation. Large-scale datasets such as AudioSet~\cite{Gemmeke2017} and VGGSound~\cite{chen2020vggsound} were primarily collected for recognition, often exhibiting weak audio-visual correspondence and lacking captions. Recent generation-oriented efforts improve data quality by filtering and captioning these datasets~\cite{zhang2024audio0synchronized,Dai_2026_CVPR}, but reduce their scale. Other datasets have limited domain coverage~\cite{ruan2022mmdiffusion,ardila2020common,afouras2018lrs3,tsuchida2019aist}, while larger multimodal collections~\cite{hochlehnert2026laion,lee2021acav100m,chen2023valor,hochlehnert2026laion} have weak audio-visual correspondence curation checks and are therefore not being used for generation. Meanwhile, large-scale training data used by frontier generative models remains proprietary. Our work addresses this gap by curating a large, high-quality, diverse set of audio-visual clips with verified correspondence and rich joint captions, supporting audio-visual generation.
\section{Curating \ourdata}

\begin{figure*}[t]
\centering
    \includegraphics[width=0.9\linewidth]{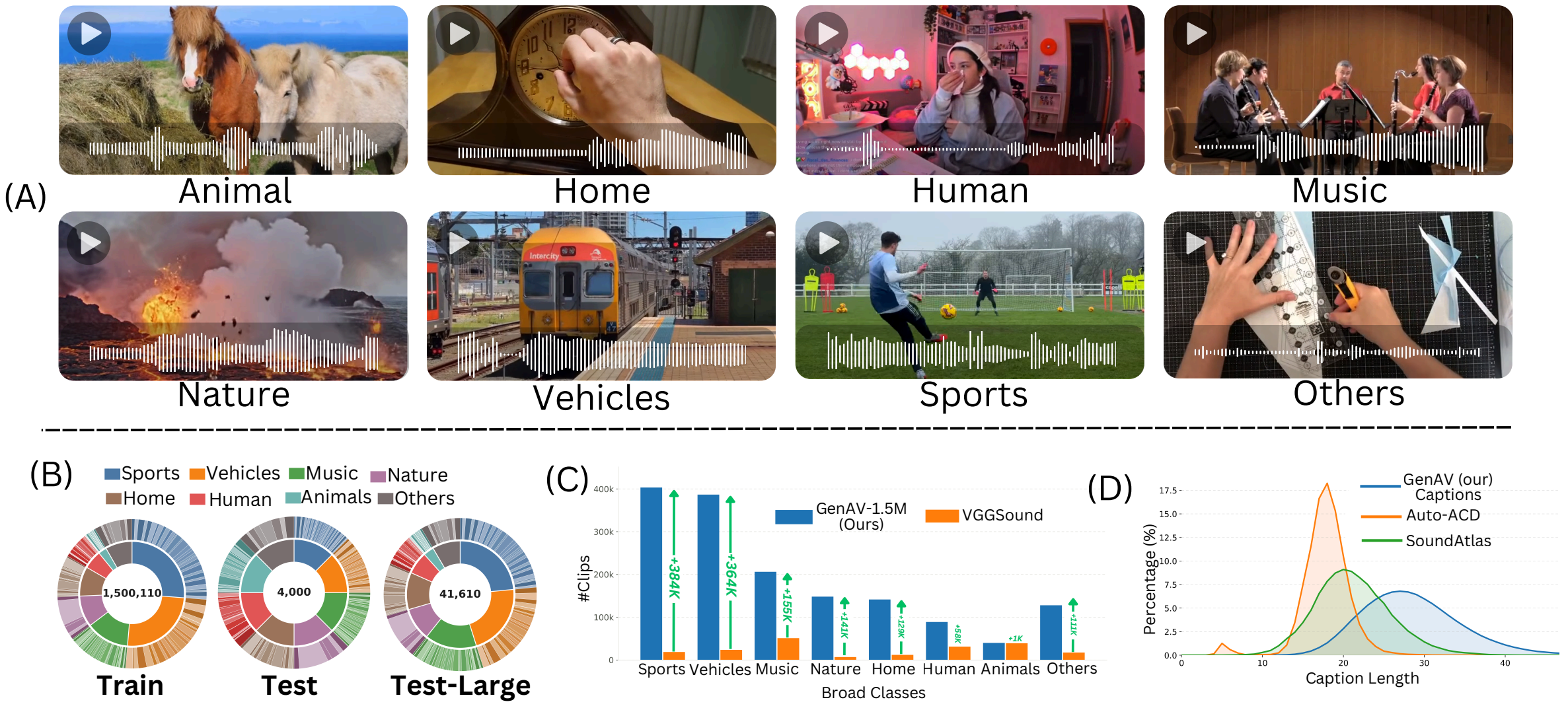}
    % \vspace{-5pt}
    \caption{(A) A representative example from \ourdata{}. (B) Class distributions across Train, Test, and Test-Large splits. (C) Comparison with VGGSound~\cite{chen2020vggsound}. (D) Caption distribution comparison with Auto-ACD~\cite{sun2024auto} and SoundAtlas~\cite{Dai_2026_CVPR}.}
    \label{fig:stats}
    % \vspace{-6mm}
\end{figure*}

As illustrated in Fig.~\ref{fig:pipeline}, our data curation pipeline consists of three stages:
(1) \textit{Data Collection and Pre-processing},
which constructs a sound-source taxonomy and retrieves and extracts candidate audio-visual clips;
(2) \textit{Filtering and Verification}, quality assessment and refinement through audio-only, video-only, and joint audio-video verification using open-source models to ensure high-quality, temporally aligned pairs
(3) \textit{Post-processing and Captioning}, which consolidates verified clips, constructs disjoint data splits, and generates rich joint audio-visual captions. \textit{Additional implementation details are in the Supplementary Material.}

\subsection{Data Collection and Pre-processing}

\noindent $\bullet$ \textbf{Sound-source Taxonomy}.
We construct a source-centric taxonomy from VGGSound~\cite{chen2020vggsound}, AudioSet~\cite{Gemmeke2017}, and Wikipedia, grouping related sound-producing events (e.g., ``dog howling'' and ``dog growling'' under ``dog'') and sources (e.g., ``owl'' and ``crow'' under ``birds''). Unlike recognition datasets, these categories serve as non-mutually-exclusive retrieval targets rather than classification labels. This yields \textbf{182 sound-source categories} across \textbf{eight groups}: \texttt{animals, home, human, music, nature, sports, vehicles,} and \texttt{others}.

\noindent $\bullet$ \textbf{Search Queries and Metadata Filtering.}
For each class, we construct $\sim$50 YouTube queries covering synonyms (e.g., ``pig" + ``piglet''), source-action pairs (e.g., ``pig oinking''), and contextual scenarios (e.g., ``pig on a farm''). Using the YouTube API, we retrieve up to 150 URLs per query, restricting results to HD videos of 4--20 minutes published within the past 15 years. We remove duplicate URLs and filter metadata associated with edited or non-natural content (e.g., ``lyrics video'' and ``animation'')~\cite{lee2021acav100m}.

\noindent $\bullet$ \textbf{Downloading and Clip Extraction.}
We use \texttt{yt-dlp} to download videos at 360p resolution, excluding unavailable videos or those without audio. A distributed pipeline manages parallel downloads across machines using PostgreSQL and shared cloud storage. To identify meaningful audio-visual events, which are often sparse within long videos, we use \texttt{PySceneDetect} to segment videos into visually coherent scenes. We discard scenes shorter than 2s and extract clips of up to 10s, sampling two clips from scenes lasting 10--30s and three from longer scenes. We additionally filter static content and slideshows to prioritize temporally informative clips.

This process yields $\sim$26M candidate clips. 
However, ensuring that these clips exhibit meaningful \textit{correspondence} between audible events and their visual sources requires a rigorous three-stage verification process, as discussed next.

\subsection{Filtering and Verification}

\noindent $\bullet$ \textbf{Audio Verification.}\label{a_verify}  We first verify whether each candidate clip from the earlier stage contains an audible event corresponding to its target sound-source class using PE$_{\textrm{AV}}$~\citep{vyas2026pushing}, a strong contrastively pretrained model with unified audio-video-text embeddings. We perform training-free, zero-shot audio-text retrieval using a group-specific vocabulary $\mathbf{T}^{G}_{aud}$, comprising detailed positive prompts for classes within group $G$ and broader descriptions for out-of-group classes. We adopt noun-verb descriptions (e.g., ``dog barking"), which we find more effective than nouns or verbs alone, and include multiple prompts per class (e.g., ``horse neighing", ``horse snorting", and ``horse galloping") to capture diverse events and reduce false negatives. On average, each $\mathbf{T}^{G}_{aud}$ contains $\sim$200 prompts.

For each prompt $p_{aud} \in \mathbf{T}^{G}_{aud}$, we construct the template \textit{``I hear \{prompt\} in this audio”} (see Appendix for prompt-selection ablations). Given an audio clip $a$ with class $c$, we rank all prompts in $\mathbf{T}^{G}_{aud}$ by cosine similarity between their text embeddings and the audio embedding. We retain the clip if at least one of the top-$5$ retrieved prompts corresponds to the target class $c$, i.e.,  $\operatorname{Top5}(a) \cap \mathbf{T}^{G,c}_{aud} \neq \emptyset$, where $\mathbf{T}^{G,c}_{aud}$ denotes the subset of prompts associated with class $c$. Top-$5$ retrieval accommodates overlapping sound events and alternative descriptions. We also human-inspect and iteratively refine $\mathbf{T}^{G}_{aud}$ for about 100 randomly sampled clips per class. The potential false positives are addressed in the subsequent, stricter audio-visual verification stage. This removes $\sim$18M clips.

\noindent $\bullet$ \textbf{Video Verification.}\label{v_verify}
We next verify whether the target sound source is visually present using zero-shot video-text retrieval with PE$_{\textrm{AV}}$. We construct group-specific vocabularies $\mathbf{T}^{G}_{vid}$ using noun-based prompts (e.g., ``cow'', ``bull'', and ``buffalo'') to capture visually diverse sources, with $\sim$278 prompts per group. Using the template \textit{``A video of \{prompt\}''}, we retain clips whose target-class prompts appear among the top-5 video-text matches. While audio verification removes $\sim$69\% of candidates, video verification rejects only $\sim$20\% of the remaining clips ($\sim$1.6M), reflecting the visual relevance of keyword-based retrieval.

\noindent $\bullet$ \textbf{Audio-Video Verification.}\label{av_verify} The previous verification stages do not explicitly establish a correspondence between audio and video. False positives may arise from spoken references to target sources, off-screen sounds, overlaid soundtracks, or incorrect matches admitted by the top-$5$ criterion. So, to jointly reason whether an audible event is grounded in a visual source, we use Qwen3-Omni~\cite{xu2025qwen3} to verify this. 

\noindent \underline{\textit{Speech and Music Suppression}}. Multimodal models such as Qwen3-Omni often struggle to identify overlapping sound events~\cite{chen2026polybench}, particularly in the presence of speech or background music. Since $\sim$72\%$^\dagger$ of our candidate clips contain at least one of these components, discarding them would substantially reduce data coverage. Instead, we selectively suppress speech and music using SAM Audio~\cite{shi2025sam}. We identify their presence from the top-$20$ audio-text retrieval results obtained during audio-only verification, requiring no additional inference. On the flagged clips, we remove speech first and subsequently music from the residual audio, if both are detected. It is to be noted that we do not suppress music from \texttt{Music} and preserve speech within the \texttt{Human} group.

\noindent \underline{\textit{Correspondence Verification.}}
We provide Qwen3-Omni with the video and corresponding audio (after suppression when applicable), using class-specific prompts refined through human inspection. Verification accounts for transient events that may fall between sampled frames (e.g., ball strikes) and naturally delayed sounds (e.g., thunder following lightning). Clips without an audible target event are rejected. This stage removes $\sim$3.35M clips.

Although Qwen3-Omni could be applied directly to all candidate clips, i.e., skipping the audio and video verifications, doing so would increase processing time by approximately $4\times {^{(\dagger)}}$ (due to its large model size and additional SAM Audio pass). Our preceding audio and video verification stages thus provide computationally efficient candidate filtering before more expensive multimodal verification.

\subsection{Post-Processing and Captioning}

\noindent $\bullet$ \textbf{Clip Expansion and Final Dataset.} Following the three verification stages as described earlier, 
we standardize the variable-length (2–10s) clips to 10s. This is consistent with prior datasets like AudioSet~\cite{Gemmeke2017} and VGGSound~\cite{chen2020vggsound}. This would ensure obtaining disjoint train and test splits.

\noindent \underline{\textit{10-second Clip Expansion}}.
We first remove substantially overlapping verified intervals and group nearby intervals whose combined duration is within 10s. Each group is then expanded symmetrically around its midpoint, adjusting for video boundaries to maintain a fixed duration. Videos with audios shorter than 10s are excluded.

\noindent \underline{\textit{Redundancy Reduction}}.
\begingroup
\renewcommand{\thefootnote}{\ensuremath{\dagger}}
To avoid over-representing individual source videos, some can go up to $\sim$138 clips/video \footnote[\value{footnote}]{Estimated using random 1,000 clips.}, we greedily select clips that maximize coverage of previously uncovered verified intervals, favoring temporally separated clips in case of ties. We retain at most 10 clips per source video and preserve the timestamps and identities of their contributing verified intervals.
\endgroup

\noindent \underline{\textit{Train-Test Split}}.
We partition source videos within each subclass into 95\% training and 5\% testing using a fixed random seed, ensuring no YouTube video appears in both splits. To further reduce test-set redundancy, we retain at most 5 clips per test video. We end up with $\sim$1.5M train clips and $\sim$41k candidate test clips.

\noindent $\bullet$ \textbf{Audio-Visual Captioning}.
To provide rich supervision at scale, we adopt a two-stage captioning pipeline that separates multimodal observation from caption generation. \\
\noindent \underline{\textit{Structured Evidence Collection}}.
We use Qwen3-Omni to extract visible actions, audible events, and their correspondence, organizing sources into \textit{audio-visual}, \textit{audio-only}, and \textit{video-only} groups. We record source attribution, salience, ambiguities, and temporally ordered speech transcripts using group-specific prompts. Unlike the earlier verification stage, captioning processes the complete 10s clips with unmodified audio. \\
\noindent \underline{\textit{Caption Synthesis}}.
We use Qwen3-30B-A3B-Instruct-2507~\cite{qwen3technicalreport} to convert structured observations into concise, factual captions (12--35 words), prioritizing salient audio-visual events while avoiding unsupported source attribution. We additionally extract normalized source labels for each correspondence group and retain them as metadata. \\
\noindent \underline{\textit{Test-Set Refinement}}.
We apply stricter correspondence verification to held-out clips, retaining salient, clearly attributable audio-visual events with minimal ambiguity.
For our primary benchmark, we sample 500 clips from each of the eight semantic groups, forming the balanced \textbf{\ourtest}. We additionally retain and provide the full refined test set of approximately 41K clips as \textbf{{\ourtest}-Large} for broader evaluation and future research.

\noindent \textbf{Human-guided filtering}. All the thresholds, hyperparameters, text vocabulary, etc. are manually verified using $\sim$100 random clips per class, using two annotators with expertise in audio-visual learning. A dashboard was created to visualize the clips, and we iteratively verified and adjusted these parameters. Using 100 clips per class gave us a balance of diversity and was also manageable to iterate and seemed to scale well. This gives us a balance of human verification and automatic scaling.
%
%%%%%%%%%%%%%%%%%%%%

\begin{table*}[t]
\centering
\renewcommand{\arraystretch}{1.1}
%%%%%%%%%%%%%%%%%%%%%%%%%%%%%%%%%%%%%%%%%%
% VIDEO-TO-AUDIO GENERATION
%%%%%%%%%%%%%%%%%%%%%%%%%%%%%%%%%%%%%%%%%%
\resizebox{0.9\textwidth}{!}{
\begin{tabular}{lcc|ccccc|ccccc}
% \toprule
\rowcolor{gray!15}
\multicolumn{13}{l}{\textit{Video-to-Audio Generation (V2A)}} \\
\midrule

\multirow{2}{*}{\bf Model} &
\multirow{2}{*}{\textbf{\textit{\#Params}}} &
\multirow{2}{*}{\textbf{\textit{\#Train}}} &
\multicolumn{5}{c}{\bf \ourtest} &
\multicolumn{5}{|c}{\bf Kling-Audio-Eval~\cite{klingfoley}} \\
\cmidrule(lr){4-8}
\cmidrule(lr){9-13}

& & &
\bf FD$_\downarrow$ &
\bf KL$_\downarrow$ &
\bf AT-Sem.$^\uparrow$ &
\bf AV-Sem.$^\uparrow$ &
\bf DeSync$_\downarrow$ &
\bf FD$_\downarrow$ &
\bf KL$_\downarrow$ &
\bf AT-Sem.$^\uparrow$ &
\bf AV-Sem.$^\uparrow$ &
\bf DeSync$_\downarrow$ \\
\midrule

FoleyCrafter~\cite{zhang2024foleycrafter}& 1.2B & 304K & 23.04 & 2.23 & 0.26 & 0.29  & 1.20 & 22.30 & 2.47 & 0.22 & 0.22 & 1.23 \\

Frieren~\cite{wang2024frieren} & 0.1B & 182K & 15.81 & 2.62 & 0.17 & 0.22 & 0.79 & 16.86 & 2.95 & 0.16 & 0.21  & 0.86 \\

MMAudio-L~\cite{cheng2025taming} & 1B & 1.13M & 13.85 & 2.16 & \textbf{0.32} & \textbf{0.35} & \underline{0.36} & 9.01 & 2.17 & \textbf{0.27} & 0.30  & 0.56 \\

Hunyuan-Foley~\cite{shan2025hunyuanvideofoleymultimodaldiffusionrepresentation} & 5.1B & \faLock & 17.21 & 2.16 & 0.25 & 0.35 & 0.44 & 9.13 & \textbf{2.08} & 0.22 & \textbf{0.34} & \underline{0.52} \\

Omni2Sound~\cite{Dai_2026_CVPR} & 1.1B & 3M
& \underline{10.71} & \underline{1.94} & 0.26 & 0.34 & 0.43
& 9.14 & 2.18 & 0.22 & \underline{0.32} & 0.59 \\

\midrule
\rowcolor{cyan!13}
\textbf{MMAudio-L}~(\textit{Ours}) & 1B & 1.5M
& \textbf{7.88} & \textbf{1.58} & \underline{0.28} & \textbf{0.35} & \textbf{0.34}
& \textbf{8.62} & \underline{2.10} & \underline{0.25} & 0.30 & \textbf{0.46} \\
\bottomrule
\end{tabular}
}

\vspace{5pt}

%%%%%%%%%%%%%%%%%%%%%%%%%%%%%%%%%%%%%%%%%%
% JOINT AUDIO-VISUAL GENERATION
%%%%%%%%%%%%%%%%%%%%%%%%%%%%%%%%%%%%%%%%%%
\resizebox{0.9\textwidth}{!}{
\begin{tabular}{lcc|cccccc|cccccc}
\rowcolor{gray!15}
\multicolumn{15}{l}{\large \textit{Joint Audio-Visual Generation (JAV)}} \\
\midrule

\multirow{2}{*}{\bf Model} &
\multirow{2}{*}{\textbf{\textit{\#Params}}} &
\multirow{2}{*}{\textbf{\textit{\#Train}}} &
\multicolumn{6}{c}{\bf \ourtest} &
\multicolumn{6}{|c}{\bf Javis-Bench-Mini~\cite{liu2025javisdit}} \\
\cmidrule(lr){4-9}
\cmidrule(lr){10-15}

& & &
\bf FVD$_\downarrow$ &
\bf FD$_\downarrow$ &
\bf AV-Sem.$^\uparrow$ &
\bf AT-Sem.$^\uparrow$ &
\bf VT-Sem.$^\uparrow$ &
\bf AV-Sync$^\uparrow$ &
\bf FVD$_\downarrow$ &
\bf FD$_\downarrow$ &
\bf AV-Sem.$^\uparrow$ &
\bf AT-Sem.$^\uparrow$ &
\bf VT-Sem.$^\uparrow$ &
\bf AV-Sync$^\uparrow$ \\
\midrule

MM-Diffusion~\cite{ruan2022mmdiffusion} & 0.4B & 22K
& 963.06 & 60.50 & 0.18 & - & - & \textbf{-0.59}
& 664.76 & 60.50 & \underline{0.18} & - & - & -0.59 \\

Ovi~\cite{low2025ovi} & 11B & \faLock & 295.41 & \underline{14.24} & \textbf{0.34} & 0.12 & \underline{0.34} & \underline{-0.60} & 283.24 & 16.97 & 0.13 & 0.10 & 0.27 & \underline{-0.59} \\

Universe-1~\cite{wang2025universe} & 6.4B & 5.5M & 114.54 & 31.20 & 0.09 & 0.10 & 0.33 & -0.60 & 218.10 & 24.42 & 0.09 & 0.08 & 0.26 & \textbf{-0.54} \\

JavisDiT++~(SFT+DPO)~\cite{liu2026javisdit++} & 2.1B & 1.13M$^\dagger$ & \underline{114.10} & 16.46 & 0.17 & \underline{0.16} & 0.33 & -0.68 & \underline{138.87} & \textbf{14.25} & 0.18 & \textbf{0.16} & \underline{0.27} & -0.64 \\

\midrule
\rowcolor{cyan!13}
\textbf{JavisDiT++~(SFT)}~(\textit{Ours}) & 2.1B & 1.5M$^\dagger$
& \textbf{70.52} & \textbf{11.33} & \underline{0.21} & \textbf{0.18} & \textbf{0.34} & -0.63
& \textbf{129.43} & \underline{16.96} & \textbf{0.19} & \underline{0.15} & \textbf{0.28} & -0.63 \\
\bottomrule
\end{tabular}
}

\vspace{5pt}
%%%%%%%%%%%%%%%%%%%%%%%%%%%%%%%%%%%%%%%%%%
% AUDIO-TO-VIDEO GENERATION
%%%%%%%%%%%%%%%%%%%%%%%%%%%%%%%%%%%%%%%%%%
\resizebox{0.9\textwidth}{!}{
\begin{tabular}{lcc|ccccc|ccccc}
% \toprule
\rowcolor{gray!15}
\multicolumn{13}{l}{\large \textit{Audio-to-Video Generation (A2V)}} \\
\midrule

\multirow{2}{*}{\bf Model} &
\multirow{2}{*}{\textbf{\textit{\#Params}}} &
\multirow{2}{*}{\textbf{\textit{\#Train}}} &
\multicolumn{5}{c}{\bf \ourtest} &
\multicolumn{5}{|c}{\bf AVSync15~\cite{zhang2024audio0synchronized}} \\
\cmidrule(lr){4-8}
\cmidrule(lr){9-13}

& & &
\bf FVD$_\downarrow$ &
\bf KVD$_\downarrow$ &
\bf AV-Sem.$^\uparrow$ &
\bf VT-Sem.$^\uparrow$ &
\bf AV-Sync$^\uparrow$ &
\bf FVD$_\downarrow$ &
\bf KVD$_\downarrow$ &
\bf AV-Sem.$^\uparrow$ &
\bf VT-Sem.$^\uparrow$ &
\bf AV-Sync$^\uparrow$ \\
\midrule

MM-Diffusion~\cite{ruan2022mmdiffusion} & 0.4B & 22K & 1036.00 & 4.40 & 0.05 & 0.07 & -0.69 & 1958.75 & 4.40 & 0.03 & 0.12 & -0.71  \\

TPoS~\cite{jeong2023power} & 1B & 179K & 487.77 & 1.66 & 0.17 & 0.30 & - & 1299.99 & 5.00 & 0.27 & 0.28 & -  \\

TempoToken~\cite{yariv2023diverse} & 1.3B & 120K & 874.12 & 7.30 & 0.12 & 0.09 & -0.66 & 1637.28 & 6.90 & 0.20 & 0.19 & -0.71  \\

AVSD~\cite{linz2024asva} & 1.1B & 1.35K
& \underline{251.08} & \underline{0.92} & 0.23 & \underline{0.32} & \underline{-0.64}
& \underline{787.15} & \underline{2.69} & 0.38 & \underline{0.29} & \textbf{-0.55} \\

Complex-A2V~\cite{kim2025semantically} & 3.1B & 40K & 376.13 & 5.76 & \underline{0.24} & 0.32 & -0.68 & 926.92 & 7.43 & \underline{0.39} & 0.28 & -0.71  \\

\midrule
\rowcolor{cyan!13}
\textbf{Wan2.1-A2V-Ctrl}~(\textit{Ours}) & 2.2B & 1.5M$^\dagger$
& \textbf{50.11} & \textbf{0.11} & \textbf{0.25} & \textbf{0.34} & \textbf{-0.63}
& \textbf{530.68} & \textbf{0.78} & \textbf{0.39} & \textbf{0.29} & \underline{-0.62} \\

\quad \textit{w/o A2V Controlnet}~\cite{wan2025}& 1.3B & \faLock
& 238.93 & 1.50 & 0.18 & 0.31 & -0.68
& 884.27 & 3.62 & 0.29 & 0.27 & -0.71 \\
\bottomrule
\end{tabular}
}
\caption{\textbf{Models trained on \ourdata{} achieve superior distributional quality across \ourtest{} and competitive generalization to external benchmarks.} We compare various open-sourced and closed-sourced models on \textit{V2A}, \textit{JAV}, and \textit{A2V} generation. To assess cross-benchmark generalization, we additionally evaluate on Kling-Audio-Eval~\cite{klingfoley} (\textit{V2A}), Javis-Bench-Mini~\cite{liu2025javisdit} (\textit{JAV}), and AVSync15~\cite{zhang2024audio0synchronized} (\textit{A2V}). $^\dagger$ denotes models with closed-source base weights adapted to \ourdata{} via LoRA~\cite{hu2021lora} or ControlNet~\cite{zhang2023adding} fine-tuning under SFT. {\faLock} indicates proprietary training data.}
\label{tab:main_table}
\end{table*}
%%%%%%%%%%%%%%%%%%%%
%
\noindent \textbf{Comparison with Existing Curation Pipelines}.
Existing audio-visual curation pipelines require adaptation to our goal of collecting large-scale, in-the-wild data for generation. Early datasets, such as AudioSet and VGGSound~\cite{chen2020vggsound,Gemmeke2017}, primarily target recognition and classification, whereas natural videos often contain multiple overlapping events rather than mutually exclusive labels~\cite{zverevwiedemer2025vggsounder}. Recent generation-oriented pipelines~\cite{Dai_2026_CVPR,zhang2024audio0synchronized} often filter previously curated datasets, which already have audio-visual correspondence. Meanwhile, advances in pretrained multimodal models have significantly improved zero-shot capabilities, enabling more flexible, open-vocabulary verification without task-specific training. We leverage these advances to develop a multi-stage pipeline built entirely on open-source models, providing a cost-effective and scalable alternative to extensive manual annotation or proprietary model APIs. Targeted human verification further guides the refinement of filtering criteria.

\begin{figure*}[t]    \centerline{\includegraphics[width=0.9\linewidth]{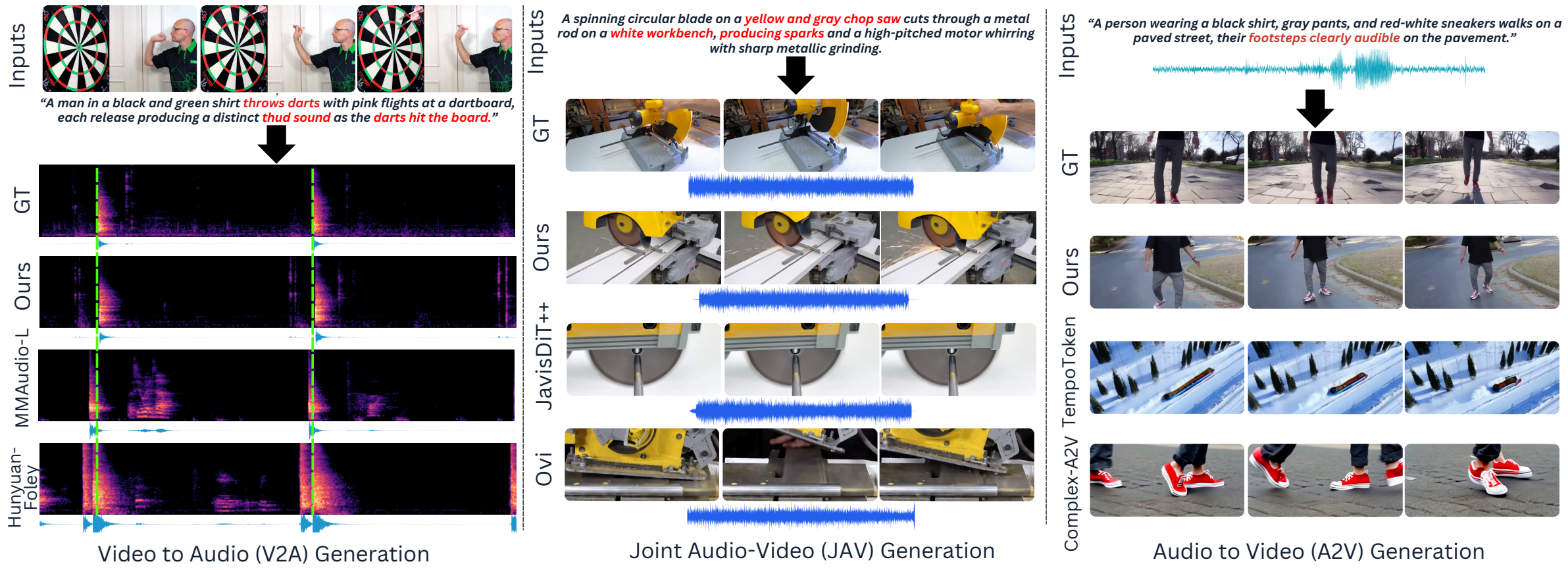}}
    \caption{We qualitatively compare our models with existing baselines and ground truth (GT).}
    \label{fig:qual_comp}
% \vspace{-5mm}
\end{figure*}

\subsection{Summary of \ourdata}

\noindent\textbf{Dataset Composition}. \ourdata{} comprises \textbf{1,541,720} audio-visual clips from 270,359 unique YouTube videos, with \textbf{1,500,110} in training, \textbf{4,000} in {\ourtest}, and \textbf{41,610} in {\ourtestLarge}. The dataset spans \textbf{eight broad class groups}, i.e., \texttt{animals, home, human, music, nature, sports, vehicles,} and \texttt{others}, further organized into \textbf{182 sound-source categories}. Fig.~\ref{fig:stats} illustrates their distributions across the train and test splits, along with representative clip frames. 
\newline
\noindent\textbf{Caption Statistics}. In \ourdata{}, we also provide rich joint audio-visual captions averaging 28.5 words, compared to 17.8 and 21.2 words as in Auto-ACD~\cite{sun2024auto} and SoundAtlas~\cite{Dai_2026_CVPR} captions on their respective VGGSound subsets. Fig.~\ref{fig:stats}(D) illustrates these distributions, highlighting the greater descriptive length of our annotations. We also release the transcriptions.
\newline
\noindent\textbf{Comparison with VGGSound}. VGGSound~\cite{chen2020vggsound} is widely used for audio-visual generation. As shown in Fig.~\ref{fig:stats}(C), \ourdata{} contains over $8\times$ more clips (1.54M vs.\ 200K), with substantially greater coverage across most semantic groups, particularly \texttt{sports}, \texttt{vehicles}, and \texttt{nature}. For a fair comparison, we map VGGSound's categories to our eight groups, merging its ``Home" and ``Tools" groups into \texttt{home}. Unlike VGGSound's taxonomy, our sound-source categories serve as flexible retrieval targets rather than mutually exclusive labels, enabling the collection of diverse sound-producing events and interactions within each category.
\section{Experiments and Results}

\subsection{Setup}

\noindent \textbf{Tasks}.
We evaluate \ourdata{} across three open-domain generation tasks:
(1) \textit{Video-to-Audio (V2A)} generates temporally aligned audio conditioned on video and text;
(2) \textit{Audio-to-Video (A2V)} generates synchronized video conditioned on audio and text; and
(3) \textit{Joint Audio-Visual (JAV)} generates semantically consistent and synchronized audio-video pairs from text. Following established evaluation settings, we use 8s clips for \textit{V2A}~\cite{cheng2025taming} and 2s clips for \textit{A2V}~\cite{yariv2023diverse} and \textit{JAV}~\cite{liu2026javisdit++}. We re-caption the 2s segments to ensure that their textual descriptions accurately reflect the shortened audio-visual content.

\noindent \textbf{Datasets}.
We compare models trained in \ourdata{} with baselines trained in public~\cite{Gemmeke2017,chen2020vggsound,zhang2024audio0synchronized} and proprietary audio-visual datasets~\cite{low2025ovi,shan2025hunyuanvideo}. We evaluate all of the baselines on GenAV-Test, which comprises 4K diverse, captioned audio-visual clips, and also on established task-specific benchmarks. For \textit{V2A}, we use Kling-Audio-Eval~\cite{klingfoley} ($\sim$21K clips with audio captions); for \textit{A2V}, AVSync15~\cite{zhang2024audio0synchronized}; and for \textit{JAV}, Javis-Bench-Mini~\cite{liu2025javisdit} ($\sim$1K clips). Additional evaluations on {\ourtestLarge} are provided in the Supplementary Material.

\noindent \textbf{Baselines}.
We compare against publicly available models trained on either open-source or closed-source audio-visual datasets. For \textit{V2A}, we evaluate FoleyCrafter~\cite{zhang2024foleycrafter}, Frieren~\cite{wang2024frieren}, MMAudio-Large-44.1~\cite{cheng2025taming}, HunyuanVideo-Foley-XXL~\cite{shan2025hunyuanvideofoleymultimodaldiffusionrepresentation}, and Omni2Sound~\cite{Dai_2026_CVPR}.
For \textit{A2V}, we compare with MM-Diffusion~\cite{ruan2022mmdiffusion} (AIST++~\cite{tsuchida2019aist} checkpoint), TPoS~\cite{jeong2023power}, TempoToken~\cite{yariv2023diverse}, AVSD~\cite{zhang2024audio0synchronized} (AVSync15 checkpoint), and Complex-A2V~\cite{kim2025semantically}. Since AVSD requires an initial frame, we provide our generated first frame. We also evaluate the Wan2.1\cite{wan2025} backbone without our audio-conditioned ControlNet~\cite{zhang2023adding} to analyze the contribution of the audio-ControlNet branch. All \textit{A2V} baselines except MM-Diffusion additionally use text conditioning. For \textit{JAV}, we compare with MM-Diffusion, Ovi~\cite{low2025ovi}, UniVerse-1~\cite{wang2025universe}, and JavisDiT++~\cite{liu2026javisdit++}. Following JavisDiT++, we provide its generated first frame to UniVerse-1. Since MM-Diffusion generates unconditionally, we sample 1,000 clips for evaluation.

\noindent \textbf{Evaluation Metrics}.
We evaluate across three axes: distributional quality, semantic alignment, and temporal synchronization. For \textit{distributional quality}, we compute the Fr\'echet Audio Distance (\textit{FD}) and KL divergence (\textit{KL}) using PANNs~\cite{kong2020panns} for audio, and Fr\'echet Video Distance (\textit{FVD}) and Kernel Video Distance (\textit{KVD}) using I3D~\cite{carreira2017quo} for video, following prior work~\cite{cheng2025taming,liu2026javisdit++,zhang2024audio0synchronized}. For \textit{semantic alignment}, we measure audio-visual (\textit{AV-Sem.}) and video-text (\textit{VT-Sem.}) similarity using ImageBind~\cite{girdhar2023imagebind}, and audio-text (\textit{AT-Sem.}) similarity using LAION-CLAP~\cite{laionclap2023}. For \textit{temporal synchronization}, we evaluate \textit{V2A} using \textit{DeSync}~\cite{iashin2024synchformer}, which measures the mean audio-video misalignment (in s) predicted by Synchformer (lower is better). For \textit{A2V} and \textit{JAV}, the generated 2s clips are shorter than Synchformer's required 4.8s input. We therefore use the \textit{AV-Sync} score~\cite{zhang2024audio0synchronized}, an unbounded synchronization classifier output, where higher values indicate stronger predicted audio-visual synchrony.

%%%%%%%%%%%%%%%%%%%%
\begin{table}[t]
\centering
\renewcommand{\arraystretch}{0.95}
\setlength{\tabcolsep}{4pt}

%%%%%%%%%%%%%%%%%%%%%%%%%%%%%%%%%%%%%%%%%%
% VIDEO-TO-AUDIO GENERATION
%%%%%%%%%%%%%%%%%%%%%%%%%%%%%%%%%%%%%%%%%%
\resizebox{0.9\linewidth}{!}{
\begin{tabular}{l|ccccc}

\rowcolor{gray!15}
\multicolumn{6}{l}{\textit{Video-to-Audio Generation (V2A)}} \\
\midrule

\textbf{Train data}
& \bf FD$_\downarrow$
& \bf KL$_\downarrow$
& \bf AV-Sem.$^\uparrow$
& \bf AT-Sem.$^\uparrow$
& \bf DeSync$_\downarrow$ \\
\midrule

VGGSound~\cite{chen2020vggsound}
& 12.15 & 2.10 & 2.80 & 0.23 & 0.38 \\
AudioSet-SA~\cite{Dai_2026_CVPR}
& 13.83 & 2.04 & 0.29 & \textbf{0.25} & \textbf{0.35} \\
\rowcolor{cyan!13}
{\ourdata}
& \textbf{8.60} & \textbf{1.55} & \textbf{0.30} & \textbf{0.25} & 0.38 \\

\bottomrule
\end{tabular}
}

\vspace{4pt}

%%%%%%%%%%%%%%%%%%%%%%%%%%%%%%%%%%%%%%%%%%
% JOINT AUDIO-VISUAL GENERATION
%%%%%%%%%%%%%%%%%%%%%%%%%%%%%%%%%%%%%%%%%%
\resizebox{0.9\linewidth}{!}{
\begin{tabular}{l|cccccc}
% \toprule
\rowcolor{gray!15}
\multicolumn{7}{l}{ \large \textit{Joint Audio-Visual Generation (JAV)}} \\
\midrule

% \multirow{2}{*}{\textbf{Train data}} &
% \multicolumn{6}{c}{\textbf{\ourtest}} \\
% \cmidrule(lr){2-7}

\textbf{Train data}
& \bf FVD$_\downarrow$
& \bf FD$_\downarrow$
& \bf AV-Sem.$^\uparrow$
& \bf AT-Sem.$^\uparrow$
& \bf VT-Sem.$^\uparrow$
& \bf AV-Sync$^\uparrow$ \\
\midrule

VGGSound~\cite{chen2020vggsound}
& 115.96 & 18.19 & 0.18 & 0.33 & 0.34 & -0.65 \\
AudioSet-SA~\cite{Dai_2026_CVPR}
& 129.64 & 17.23 & \textbf{0.22} & \textbf{0.37} & 0.34 & -0.65 \\
\rowcolor{cyan!13}
{\ourdata}
& \textbf{76.67} & \textbf{11.95} & 0.21 & 0.31 & \textbf{0.35} & \textbf{-0.63} \\

\bottomrule
\end{tabular}
}

\vspace{4pt}

%%%%%%%%%%%%%%%%%%%%%%%%%%%%%%%%%%%%%%%%%%
% AUDIO-TO-VIDEO GENERATION
%%%%%%%%%%%%%%%%%%%%%%%%%%%%%%%%%%%%%%%%%%
\resizebox{0.9\linewidth}{!}{
\begin{tabular}{l|ccccc}
% \toprule
\rowcolor{gray!15}
\multicolumn{6}{l}{\textit{Audio-to-Video Generation (A2V)}} \\
\midrule

\textbf{Train data}
& \bf FVD$_\downarrow$
& \bf KVD$_\downarrow$
& \bf AV-Sem.$^\uparrow$
& \bf VT-Sem.$^\uparrow$
& \bf AV-Sync$^\uparrow$ \\
\midrule

VGGSound~\cite{chen2020vggsound}
& 72.76 & 0.30 & 0.25 & 0.33 & -0.63 \\
AudioSet-SA~\cite{Dai_2026_CVPR}
& 77.93 & 0.45 & 0.24 & 0.33 & -0.64 \\
\rowcolor{cyan!13}
{\ourdata}
& \textbf{50.78} & \textbf{0.12} & \textbf{0.25} & \textbf{0.34} & \text{-0.63} \\

\bottomrule
\end{tabular}
}

\caption{
\textbf{\ourdata{} proves to be an effective training dataset across audio-visual generation tasks.} We compare models trained on VGGSound, AudioSet-SA, and {\ourdata} evaluated on {\ourtest}. All comparisons use task-specific evaluation metrics.}
\label{tab:ablation}
% \vspace{-5mm}
\end{table}
%%%%%%%%%%%%%%%%%%%%
\subsection{Results and Discussions}

Tab.~\ref{tab:main_table} compares our models with existing baselines. \textit{Implementation details are provided in the supplementary.}

\noindent \textbf{Improvements across generation tasks}.
Models trained on {\ourdata} achieve the best distributional quality across all tasks on {\ourtest}, with strong performance on external benchmarks. Improvements are substantial for video generation, reducing \textit{FVD} from 114.10 to 70.52 for \textit{JAV} and 251.08 to 50.11 for \textit{A2V} over the strongest baselines. Beyond distributional quality, our models achieve strong synchronization (best \textit{DeSync} for \textit{V2A}) and semantic alignment scores (best \textit{AV-Sem.} and \textit{VT-Sem.} for \textit{A2V}). These results demonstrate the effectiveness of {\ourdata} across diverse generative tasks and evaluation dimensions.

\noindent \textbf{Generalization to External Benchmarks}.
Our models demonstrate strong generalization across unseen benchmarks, achieving the lowest \textit{FD} (8.62) and \textit{DeSync} (0.46) on Kling-Audio-Eval, the lowest \textit{FVD} (129.43) on Javis-Bench-Mini, and consistently better on AVSync15. This indicates that training models on high-quality and diverse coverage of {\ourdata} can consistently generalize across other established audio-visual generation datasets.

\noindent \textbf{Improving Existing Generative Models}.
Training on {\ourdata} improves architectures without increasing model capacity. Our MMAudio-L reduces \textit{FD} from 13.85 to 7.88 on {\ourtest}, using only audio-video-text pairs, without the additional audio-text data or data balancing employed in the original training. Similarly, our SFT-only JavisDiT++ achieves better or competitive performance than the published SFT+DPO checkpoint, highlighting the effectiveness of {\ourdata} without additional DPO.

\noindent \textbf{Competitive with Larger Proprietary-Data Models}.
Our models achieve competitive performance against substantially larger models trained on proprietary data. On Kling-Audio-Eval, our 1.03B MMAudio-L outperforms the 5.13B HunyuanVideo-Foley in \textit{FD} (8.62 vs.\ 9.13) and\textit{ DeSync} (0.46 vs.\ 0.52). Similarly, our 2.1B JavisDiT++ achieves lower \textit{FVD} than the 11B Ovi on Javis-Bench-Mini (129.43 vs.\ 283.24). These results highlight the potential and need for high-quality, accessible AV training data.

\noindent \textbf{Qualitative Comparison}.
Fig.~\ref{fig:qual_comp} presents qualitative comparisons. Our models capture action-aligned sound events in \textit{V2A}, generate coherent audio-visual content from text in \textit{JAV}, and produce visually plausible motion corresponding to the input audio in \textit{A2V}. These examples further illustrate the benefits of {\ourdata} for generation quality and cross-modal correspondence.

\begin{figure}[t]    \centerline{\includegraphics[width=0.9\linewidth]{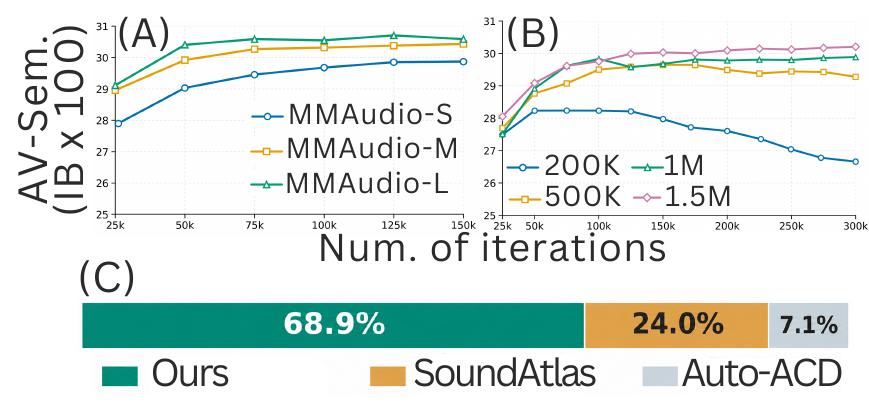}}
    % \vspace{-2mm}
    \caption{\textbf{(A) and (B) Scaling dataset size and model parameters improves the audio-visual correspondence.} In (A), we train three variants of MMAudio and evaluate on \ourtest{}. In (B), we train MMAudio on different increasing dataset subsets. \textbf{(C) Humans prefer ``Ours" over SoundAtlas and Auto-ACD.}}
    % \vspace{-5mm}
    \label{fig:add_exp}
\end{figure}

\subsection{Additional Analysis}

\noindent \textbf{Controlled Comparison of Training Data}.
We compare training datasets while keeping everything else fixed. We consider VGGSound~\cite{chen2020vggsound} ($\sim$180K), AudioSet-SA ($\sim$232K), an AudioSet subset filtered for audio-visual correspondence by SoundAtlas~\cite{Dai_2026_CVPR}, and {\ourdata} ($\sim$1.5M). For \textit{V2A}, we train MMAudio-S for 100 epochs; for \textit{JAV}, we fine-tune JavisDiT++ using SFT for one epoch; and for \textit{A2V}, we train the Wan2.1-based audio-conditioned ControlNet from scratch for one epoch. As shown in Tab.~\ref{tab:ablation}, {\ourdata} achieves the best distributional quality across all three tasks, demonstrating its effectiveness as a general-purpose training resource for audio-visual generation.

\noindent \textbf{Scaling with Dataset Size}.
In this section, we study the effect of training data size on audio-visual correspondence. We construct randomly selected and group-class stratified subsets of 200K, 500K, 1M, and full {\ourdata} train clips. We train MMAudio-S (16\,kHz) on each subset and evaluate \textit{AV-Sem.} on {\ourtest} throughout training. As shown in Fig.~\ref{fig:add_exp}(B), performance improves with increasing data size, with the full dataset achieving the highest correspondence. Smaller subsets exhibit early saturation and even degradation with continued training, whereas larger subsets maintain higher scores. These results suggest that additional {\ourdata} provides useful training signals rather than merely redundant examples.

\noindent \textbf{Scaling with Model Size}.
We further investigate whether increasing model capacity improves audio-visual correspondence when trained on {\ourdata}. We train three \textit{V2A} MMAudio variants, i.e., Small (157M), Medium (622M), and Large (1.03B), and evaluate \textit{AV-Sem.} on {\ourtest} throughout training. As shown in Fig.~\ref{fig:add_exp}(A), performance consistently improves with model size, with the largest model achieving the highest correspondence. These results demonstrate that {\ourdata} supports effective scaling to generative models.

\noindent \textbf{Caption Quality}.
To evaluate our captioning pipeline, we compare our generated captions with those generated by Auto-ACD~\cite{sun2024auto} and SoundAtlas~\cite{Dai_2026_CVPR} on 20 randomly sampled VGGSound clips. As shown in Fig.~\ref{fig:stats}, our captions are more descriptive, with a higher average word count than both baselines. We further conduct a human preference study with 17 participants, who are presented with each video and three candidate captions and asked: \textit{``Which caption best aligns with both audio and video?''} GenAV captions are preferred in \textbf{68.9\%} of judgments, compared to 24.0\% for SoundAtlas and 7.1\% for AutoACD (Fig.~\ref{fig:add_exp}(C)). 
These results demonstrate a strong human preference for our captions based on audio-visual alignment.
\section{Conclusion}

We introduced \textbf{GenAV-1.5M}, a large-scale, in-the-wild audio-visual dataset comprising 1.54 million clips across 182 sound-source categories, with rich joint audio-visual captions. To construct this dataset, we developed a scalable, cost-effective curation pipeline that combines zero-shot unimodal filtering, selective audio suppression, and joint audio-visual correspondence verification using open-source models. Extensive experiments across video-to-audio, audio-to-video, and joint audio-visual generation demonstrate the effectiveness of GenAV-1.5M in improving generation quality and generalizing to external benchmarks. \\

\noindent \textbf{Acknowledgments.} This work was supported in part by an Amazon Research Award. We thank High Performance Computing at The University of Texas at Dallas (HPC@UTD) for providing computing resources and support. We also thank Aditya Kulkarni for his help in exploring existing datasets and Tushar Goyal for his assistance with preliminary experiments. Any opinions, findings, and conclusions expressed in this material are those of the authors and do not necessarily reflect the views of Amazon or HPC@UTD.

% \clearpage

{
    \small
    \bibliographystyle{ieeenat_fullname}
    \bibliography{main}
}

\clearpage
\maketitlesupplementary

\addtocontents{toc}{\protect\setcounter{tocdepth}{2}}
\setcounter{tocdepth}{2}

\hypersetup{linkcolor=black}
\tableofcontents

\section{More details about curating \ourdata}

\noindent We provide additional details about the curation pipeline.

\subsection{Data Collection and Pre-processing}

\noindent $\bullet$ \textbf{Sound-source taxonomy}.
We start by constructing a taxonomy of sound-producing sources and interactions that are both audible and visually identifiable, drawing primarily from VGGSound and AudioSet, with Wikipedia used to identify related and hierarchical sources. This yields \textbf{182 sound-source classes} organized into \textbf{eight broad} categories:~\texttt{animals, home, human, music, nature, sports, vehicles,} and \texttt{others}.
Our group and their fine-grained classes are shown in Tab.~\ref{supp_tab:class_names_avgen}
%%%%%%%%%%%%%%%%%%%%%%%%%%
\begin{table*}[t]
\centering
\renewcommand{\arraystretch}{1.1}
\setlength{\tabcolsep}{5pt}

\begin{tabularx}{\textwidth}{lX}
\toprule
\textbf{Group} & \textbf{Sound sources} \\
\midrule

\textit{NATURE}
& Fire, Rain, Thunderstorm, Volcanic Eruption, Water, Wind \\
\midrule

\textit{MUSIC}
& Accordion, Bagpipes, Banjo, Bassoon, Bell, Bowed\_String, Clarinet, Didgeridoo, Disc\_Scratching, Double\_Bass, Drum, Flute, French\_Horn, Gong, Guitar, Harmonica, Harp, Mallet Percussion, Mandolin, Oboe, Orchestra, Organ, Piano, Saxophone, Singing Bowl, Sitar, Synthesizer, Tambourine, Theremin, Trombone, Trumpet, Tuning\_Fork, Ukulele, Violin, Wind\_Chime, Zither  \\
\midrule

\textit{VEHICLE}
& Acceleration+Revving, Airplane, Bicycle, Car\_Passing\_By, Car+Racecar, Drone, Engine\_Knocking, Engine\_Starting, Helicopter, Idling, Lawn\_Mower, Motorboat+Speedboat, Motorcycle, Opening\_Door+Closing\_Door, Power\_Window+Electric\_Window, Reversing\_Beeps, Rowboat+Kayak+Canoe, Sailboat, Ship, Sirens, Skateboard, Skidding, Snowmobile, Subway+Metro+Underground, Train, Truck+Bus, Vehicle\_Horn, Vroom   \\
\midrule

\textit{SPORTS}
& American\_Football, Arrow+Darts, Autoracing, Badminton, Baseball, Basketball, Bowling, Boxing, Cricket, Cycling\_Bmx, Field\_Hockey, Golf, Gymnastics, Handball, Ice\_Hockey, Lacrosse, Motocross, Padel+Pickleball, Parkour, Pool+Billiards, Rodeo+Equestrian, Rope Skipping, Rowing+Canoe+Kayak, Skating, Skiing, Soccer, Squash, Surfing, Swimming, Table Tennis, Track and Field, Trampoline, Volleyball, Weightlifting  \\
\midrule

\textit{HUMAN} 
& Bite+Chew, Blowing\_Nose, Breathing, Burp, Coughing, Crowd\_Cheering, Crying, Dance, Fart, Female\_Speech, Footsteps, Gargle, Hiccup, Hum, Jump, Laughter, Male\_Speech, Shouting, Sigh, Singing, Snap+Clap+Slap, Sneezing, Throat\_Clearing, Wheeze\_Pant, Whisper, Whistle  \\
\midrule

\textit{HOME} 
& Air\_Conditioning\_Machine, Camera, Clock+Alarm, Door+Drawers, Drills, Electric\_Razor+Shaver, Fan, Food\_Prep, Hammer, Mouse\_Clicking, Printer, Scissors+Knife, Sewing\_Machine, Telephone, Toilet\_Flush, Typing, Vacuum, Washing\_Machine, Woodworking, Writing  \\
\midrule

\textit{ANIMALS} 
& Amphibians\_Frogs, Aquatic\_Audible, Big\_Cat, Birds\_of\_Prey, Canid, Caprine\_Ovine, Cat, Cattle, Dinosaurs, Dog, Duck, Elephant, Equid, Forest\_Birds, Goose, Horse, Insects, Monkey, Penguins, Pig, Poultry, Reptiles, Turkey, Urban\_Birds  \\
\midrule

\textit{OTHERS} 
& Fireworks, Firing\_Cannon, Gun, Impact, Liquid, Material\_Change, Missile\_Launch, Surface\_Interaction  \\

\bottomrule
\end{tabularx}
\caption{List of groups and sound sources in {\ourdata} dataset. Some similar sound sources were combined and use ``+" symbol between them. }
\label{supp_tab:class_names_avgen}
\end{table*}
%%%%%%%%%%%%%%%%%%%%%%%%%%

\noindent $\bullet$ \textbf{Search queries, URL retrieval, and filtering.}
For each sound-source class, we construct $\sim$50 diverse YouTube search queries to retrieve videos featuring natural sound-producing objects and events. For example, some search queries for the big\_cat category are shown in Fig.~\ref{fig:query_example}

\begin{figure}[h]
\centering
\begin{tcolorbox}[
    colback=gray!5,
    colframe=gray!35,
    boxrule=0.4pt,
    arc=1mm,
    left=5pt, right=5pt,
    top=4pt, bottom=4pt,
    width=\columnwidth
]
\footnotesize
\begin{tabular}{@{}p{0.25\linewidth}p{0.69\linewidth}@{}}
\textbf{Type} & \textbf{Search Keywords} \\
\midrule
\texttt{core} &
big cat, wild cat, large feline \\[2pt]
\texttt{species} &
lion, tiger, leopard, jaguar, cheetah \\[2pt]
\texttt{vocalizations} &
lion roaring, tiger chuffing, tiger roaring, cougar screaming \\
\end{tabular}
\end{tcolorbox}
\vspace{-6pt}
\caption{\textbf{Example search queries} for the \texttt{big\_cat} sound-source category.}
\label{fig:query_example}
\end{figure}

\begin{figure}[h] 
\centerline{\includegraphics[width=0.95\linewidth]{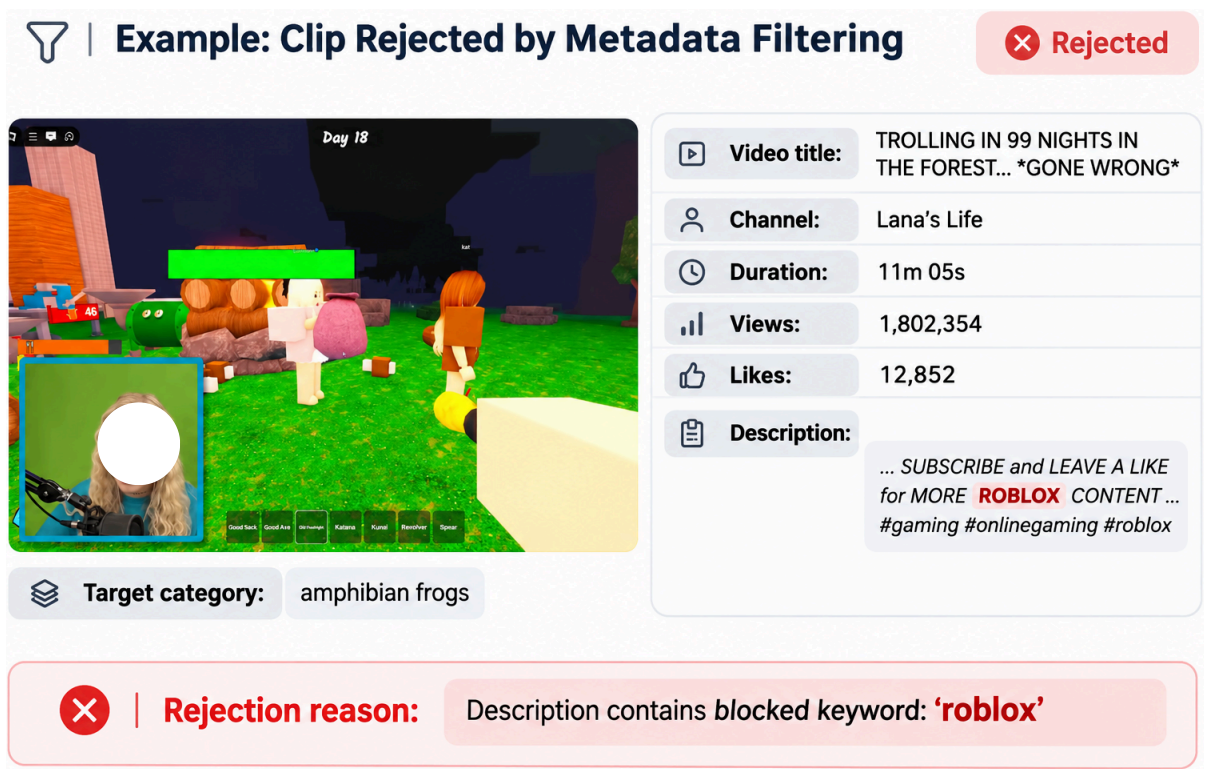}}
    \caption{Example for rejected clip during metadata filtering.}
    \vspace{-1mm}
    \label{fig:meta_rej}
\end{figure}

We restrict retrieval to \texttt{HD}, medium-duration (4--20 minutes) videos published within the past 15 years. Short videos often contain edited content, overlaid music, or synthetic effects, while longer videos incur greater downloading and processing costs.
We further filter video titles and descriptions using negative regex patterns to exclude irrelevant or artificially edited content (e.g., music videos, lyrics videos, and animations), as illustrated in Tab.~\ref{tab:metadata_filter} and Fig.~\ref{fig:meta_rej}.

\begin{table}[t]
\centering
\vspace{-3pt}
\footnotesize
\setlength{\tabcolsep}{4pt}
\begin{tabular}{@{}p{0.17\columnwidth}p{0.76\columnwidth}@{}}
\toprule
\textbf{Class} & \textbf{Excluded Keywords/Phrases} \\
\midrule
Global & Cartoon, Minecraft, stuffed animal, official music video, lyrics \\[2pt]
Goat & Messi (\textit{GOAT acronym}), sleep (\textit{sleep-aid videos}) \\[2pt]
Cattle & beef, steak, BBQ, grill, burger  (\textit{food}) \\
\bottomrule
\end{tabular}
\caption{\textbf{Examples of negative metadata filters} for the \texttt{animals} group, including global and category-specific exclusion terms.}
\label{tab:metadata_filter}
\end{table}

\noindent $\bullet$ \textbf{Download and Clip Extraction.}
To optimize download speed, storage, and processing efficiency, we use \texttt{yt-dlp} to download videos at 360p resolution. We discard videos with missing audio or unavailable URLs. As shown in Fig.~\ref{fig:add_exp}, to scale retrieval, we implement a distributed downloading pipeline using PostgreSQL to partition URLs across multiple machines, each downloading videos in parallel and uploading them to shared cloud storage while periodically synchronizing the central database. 

\begin{figure}[h] 
\centerline{\includegraphics[width=0.95\linewidth]{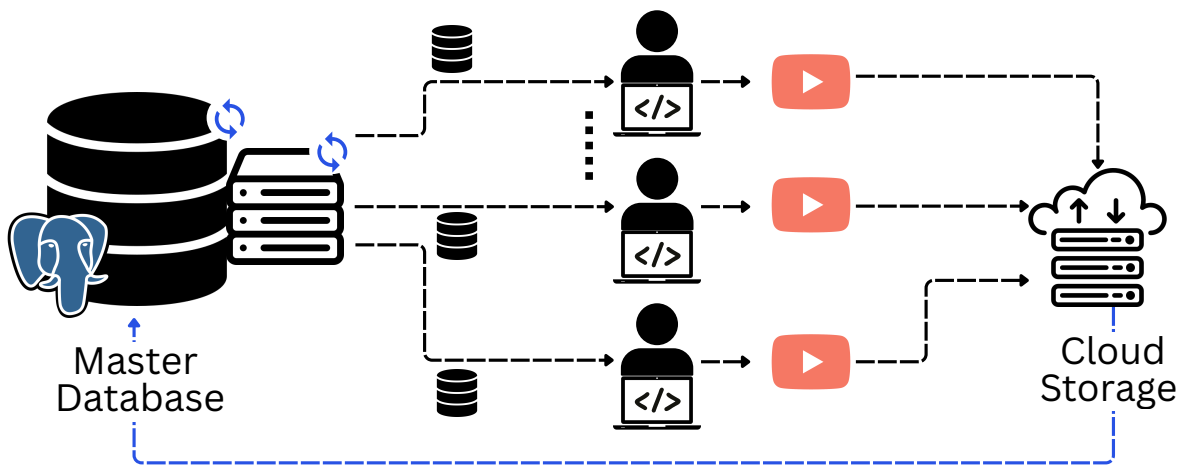}}
    \caption{Our distributed download pipeline.}
    \vspace{-1mm}
    \label{fig:add_exp}
\end{figure}

Since meaningful audio-visual occurrences are often sparse within a video, we employ \texttt{PySceneDetect} to split videos into visually coherent scenes. We discard scenes shorter than 2s and extract clips of \textit{up to 10s}. We sample two clips (i.e., the first 10s and the last 10s) if a video lasts 10--30s and three (i.e., first 10s, middle 10s, and last 10s) from scenes exceeding 30s, each of 10s. We further extend \texttt{PySceneDetect} to detect and filter static content and slideshows, favoring temporally rich videos, which is critical for audio-visual learning tasks like generation.

For each video, we use \texttt{PySceneDetect} to compute an inter-frame difference score $d_t$, defined as the mean absolute pixel-wise difference between frames $t-1$ and $t$ in HSV color space, where $t$ denotes the frame index. 
We discard videos with a median inter-frame difference 
$\tilde{d}=\mathrm{median}_t\,d_t < 0.5$. 
We use the median rather than the mean so that a few abrupt
changes (e.g., slide transitions) cannot make a mostly frozen video look dynamic. 
Among the discarded videos, we distinguish between static videos and slideshows based on the maximum inter-frame 
difference $d_{\max}=\max_t d_t$, classifying those with $d_{\max}<10$ as static videos and those with $d_{\max}\geq10$ as slideshows. Both thresholds are empirically determined through manual inspection of diverse video samples.

\subsection{Filtering and Verification}

\noindent $\bullet$ \textbf{Audio Verification}.
Following prior zero-shot audio classification approaches~\cite{elizalde2023clap,kushwaha2023multimodal}, we evaluate different audio-text prompt templates on ESC-50~\cite{piczak2015dataset}. As shown in Tab.~\ref{tab:prompt_ablation}, we adopt \textit{``I hear \{prompt\} in this audio''}, which naturally accommodates noun-verb descriptions (e.g., ``dog barking'' or ``horse neighing'') while achieving high top-1 accuracy (93.85\%), outperforming the bare-label baseline (89.55\%) by 4.30\%. We use this template for all audio prompts. \\

\begin{table}[t]
\centering
\footnotesize
\setlength{\tabcolsep}{4pt}
\renewcommand{\arraystretch}{1.05}
\begin{tabularx}{\columnwidth}{@{}Xr@{}}
\toprule
 \textbf{Prompt Template} & \textbf{Top-1 Acc.} \\
\midrule
``This audio contains \{\} sounds.'' & 93.85\% \\
\textbf{``I hear \{\} in this audio.''} & \textbf{93.85}\% \\
``a sound of \{\}'' & 93.70\% \\
``This audio segment features \{\}.'' & 93.35\% \\
``The audio contains the sound of \{\}.'' & 93.10\% \\
``The audio is classified as \{\}.'' & 92.95\% \\
``This is a sound of a \{\}.'' & 92.75\% \\
``The sound in this clip is \{\}.'' & 92.70\% \\
``a recording of \{\}'' & 92.40\% \\
``This is a recording of \{\}.'' & 90.90\% \\
``\{\}'' & 89.55\% \\
\bottomrule
\end{tabularx}
\vspace{-3mm}
\caption{\textbf{Audio-text prompt template ablation on ESC-50.} Top-1 classification accuracy using different text templates.}
\label{tab:prompt_ablation}
\vspace{-8mm}
\end{table}

% \vspace{-2mm}
\noindent $\bullet$\textbf{Video Verification.}\label{v_verify} After the audio-only verification, we also assess whether each clip visually depicts its target sound source using PE$_{\textrm{AV}}$~\cite{vyas2026pushing}. Similar to audio verification, we construct group-specific video-text vocabularies $\mathbf{T}^{G}_{vid}$, but employ simpler noun-based prompts, as visual appearances often provide clearer cues for distinguishing acoustically similar sources (e.g., rainfall and food frying). To include visual diversity within a sound-source, we include multiple related sources (e.g., ``cow", ``bull", ``buffalo", and ``yak"), yielding $\sim$278 prompts per group, compared to 200 for audio. 
As the video-text template, we use \textit{``A video of \{prompt\}''}. For computational efficiency, we uniformly sample 64 frames from each clip as input to the PE$_{\textrm{AV}}$ video encoder. Similar to audio verification, we pre-compute the text embeddings and perform zero-shot video-text retrieval, retaining clips if at least one target-class prompt appears among the top-$5$ matches.
%
%%%%%%%%%%%%%%%%%%%%%%%%%%
\begin{lstlisting}[
    caption={Example audio-visual verification prompt for the Animals group.},
    label={lst:av_prompt},
    captionpos=b,
    basicstyle=\ttfamily\scriptsize,
    backgroundcolor=\color{gray!5},
    frame=single,
    breaklines=true,
    columns=fullflexible,
    linewidth=\columnwidth
]
Given a video clip and its speech-suppressed audio, verify whether [target instrument] is audibly and visually aligned. Retrieval predictions are not evidence.

Audio: Independently determine whether the target instrument is audible, including quiet or brief passages. Do not infer sound from visibility or confuse speech, noise, or artifacts with the target.

Video: Determine whether the instrument is visible and actively played. Merely holding or displaying it is insufficient.

Correspondence: Check whether the sound matches the visible playing action in source, rhythm, intensity, and timing. A brief matching passage is sufficient.

Verdict:
ALIGNED: All checks pass.
MISALIGNED: Target is audible but lacks a matching visible source.
NO_TARGET_SOUND: Otherwise.

Output: Return JSON containing target class, verdict, visibility/audibility/correspondence booleans, separate visual and auditory evidence, justification, and a visual-only scene description.
\end{lstlisting}
%%%%%%%%%%%%%%%%%%%%%%%%%%

Notably, we observe that while audio-only verification eliminates $\sim$54\% ($\sim$18M) of candidate clips, video-only verification rejects only 20\%($\sim$1.6M) of the remaining clips, reflecting the visual relevance of keyword-based YouTube retrieval. \\

\noindent $\bullet$ \textbf{Audio-Video Verification.}\label{av_verify}

\noindent \underline{\textit{Speech and Music Suppression}}.
We use pretrained SAM-Audio (\texttt{facebook/sam-audio-large-tv}) to selectively suppress speech and music in candidate clips based on upstream detection flags. Audio is decoded to 48\,kHz mono, and silent clips (RMS $<10^{-4}$) are skipped. We use the prompts ``speech, single speaker'' and ``music'' for separation. When both are detected, we remove speech first, then remove music from the residual audio. Inference uses four reranking candidates with only text prompts (i.e., no span or video prompts). Clips are processed in batches and with multiple data-loading workers to make the processing faster.

\begin{figure}[h]
    \centering
    \includegraphics[width=0.95\linewidth]{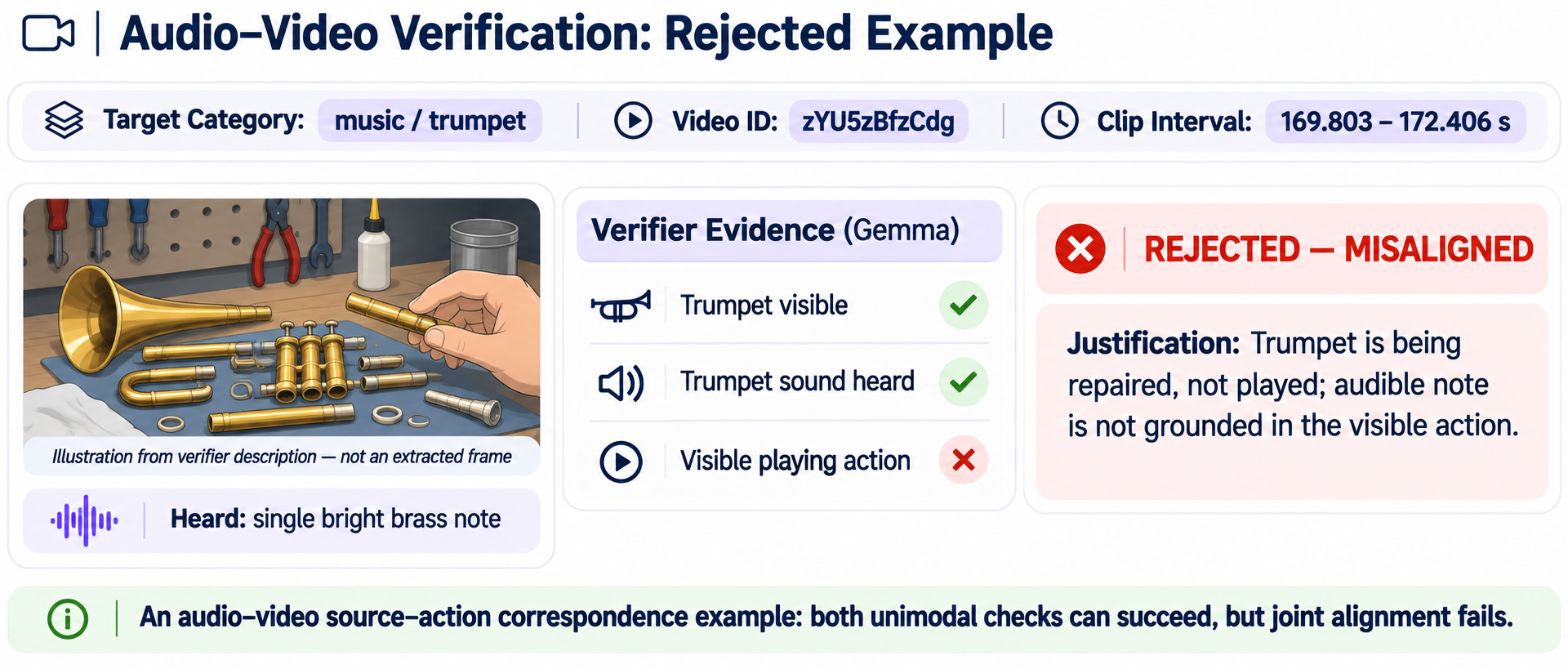}
    \caption{\textbf{Audio-visual correspondence verification.}
    An example of a rejected clip where the target instrument and sound are present, but the visible action does not explain the audible event.}
    \vspace{-1mm}
    \label{fig:av_verify}
\end{figure}

\noindent \underline{\textit{Correspondence Verification.}}
The preceding audio and video verification stages establish source-presence independently but do not ensure correspondence between audible events and visible sources. For instance, an animal may be visible while its calls originate from an off-screen source, or unrelated sounds may be overlaid on the video. We therefore use Qwen3-Omni~\cite{xu2025qwen3} with group-specific prompts to independently assess audible events, visible source activity, and their correspondence. Listing~\ref{lst:av_prompt} provides a representative verification prompt for the Animals group. An example is also illustrated in Fig.~\ref{fig:av_verify}.

\subsection{Post-Processing and Captioning}

\noindent \textbf{Audio-Visual Captioning: Additional Details}

\noindent \underline{Structured Evidence Collection}.
We use Qwen3-Omni-30B-A3B-Thinking to jointly process the complete 10-second video and its original, unmodified audio. The model generates structured observations of visible actions, audible events, and their correspondence, separating sources into \textit{audio-visual}, \textit{audio-only}, and \textit{video-only} groups. For each audio-visual event, we record source visibility and activity, acoustic characteristics, sound salience, event prominence, association strength, and attribution ambiguity. We additionally record speech, music, scene transitions, and potential confounders, with intelligible speech transcribed in temporal order. Group-specific prompts guide evidence collection without using the retrieved subclass or filename as evidence.

\noindent \underline{Caption Synthesis}.
We pass the structured observations to Qwen3-30B-A3B-Instruct-2507 to generate concise, factual joint captions, preferably 12--35 words. Captions prioritize salient visible actions and their corresponding sounds, while incorporating secondary events when relevant. The model is instructed to avoid unsupported source attribution, temporal relationships, and details not grounded in the observations. Speech transcripts are excluded from the caption-generation input and retained separately to prevent unnecessary reproduction of spoken content. We also extract normalized source labels for the three correspondence groups, consolidating redundant descriptions of the same physical source.

\noindent \underline{Test-Set Refinement}.
We apply an additional taxonomy-guided selection stage to held-out clips using Qwen3-30B-A3B-Instruct-2507. Candidate selections undergo rule-based validation of source attribution, subclass relevance, event prominence, sound clarity, ambiguity, and media policy. We incorporate broad-class specific exceptions rather than imposing uniform criteria across all sound sources. Sports clips additionally undergo audio-only verification, with targeted retries and confirmation checks for ambiguous cases. These stricter checks are applied only to the test set.

\begin{figure}[h]
    \centering
\includegraphics[width=0.98\linewidth]{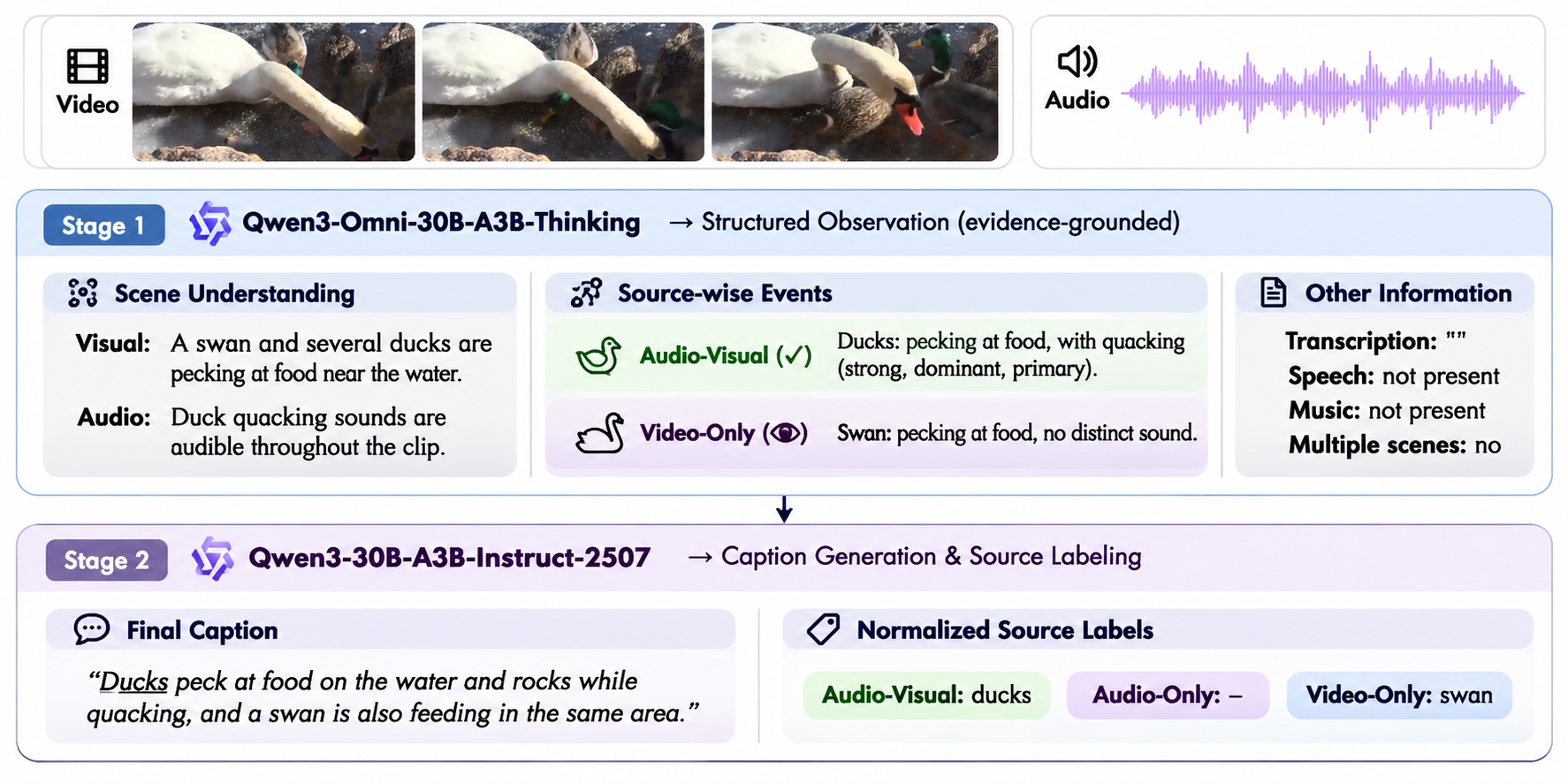}
    \caption{\textbf{Example of our two-stage audio-visual captioning pipeline.}
    Qwen3-Omni extracts structured observations distinguishing audio-visual, audio-only, and video-only sources. Qwen3-Instruct synthesizes a grounded joint caption and normalized source labels. The example is illustrative rather than an actual model output.}
    \vspace{-4mm}
    \label{fig:caption}
\end{figure}

\begin{figure}[h]
    \centering
\includegraphics[width=0.9\linewidth]{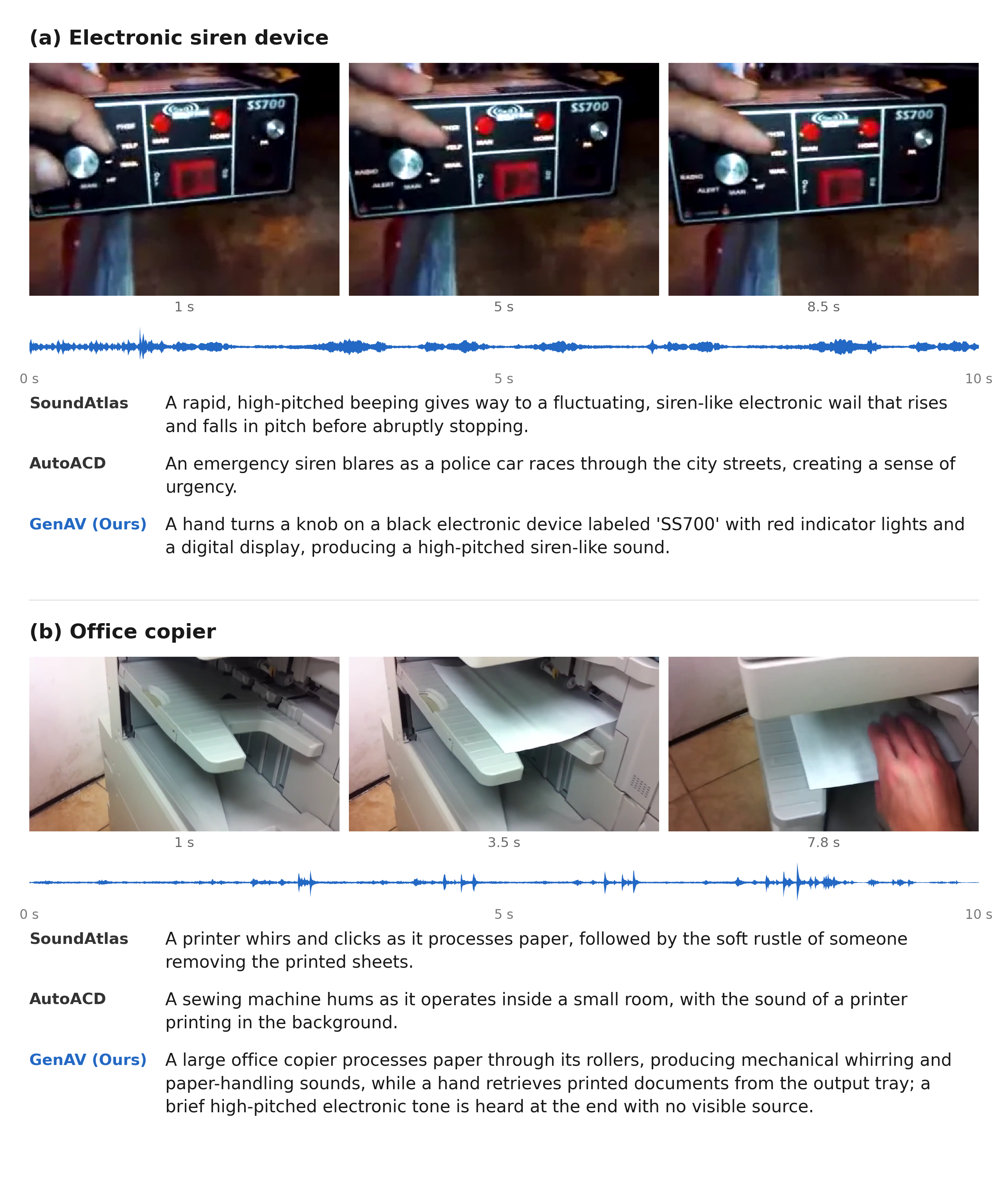}
    \caption{\textbf{Caption Quality Comparison}. Comparing caption from SoundAtlas, AutoACD and Ours. We observe that our captions are more descriptive and aligns more with audio and video.}
    \vspace{-1mm}
    \label{fig:caption_example}
\end{figure}

The captioning pipeline is illustrated in Fig.~\ref{fig:caption}. We also show two examples of our captioning pipeline (compared with SoundAtlas and AutoACD) in Fig.~\ref{fig:caption_example}.
%%%%%%%%%%%%%%%%%%%%%%%%%%%%%%%%%
%%%%%%%%%%%%%%%%%%%%%%%%%%%%%%%%%
%%%%%%%%%%%%%%%%%%%%%%%%%%
%%%%%%%%%%%%%%%%%%%%%%%%%%

\begin{figure}[h]
    \centering
\includegraphics[width=0.95\linewidth]{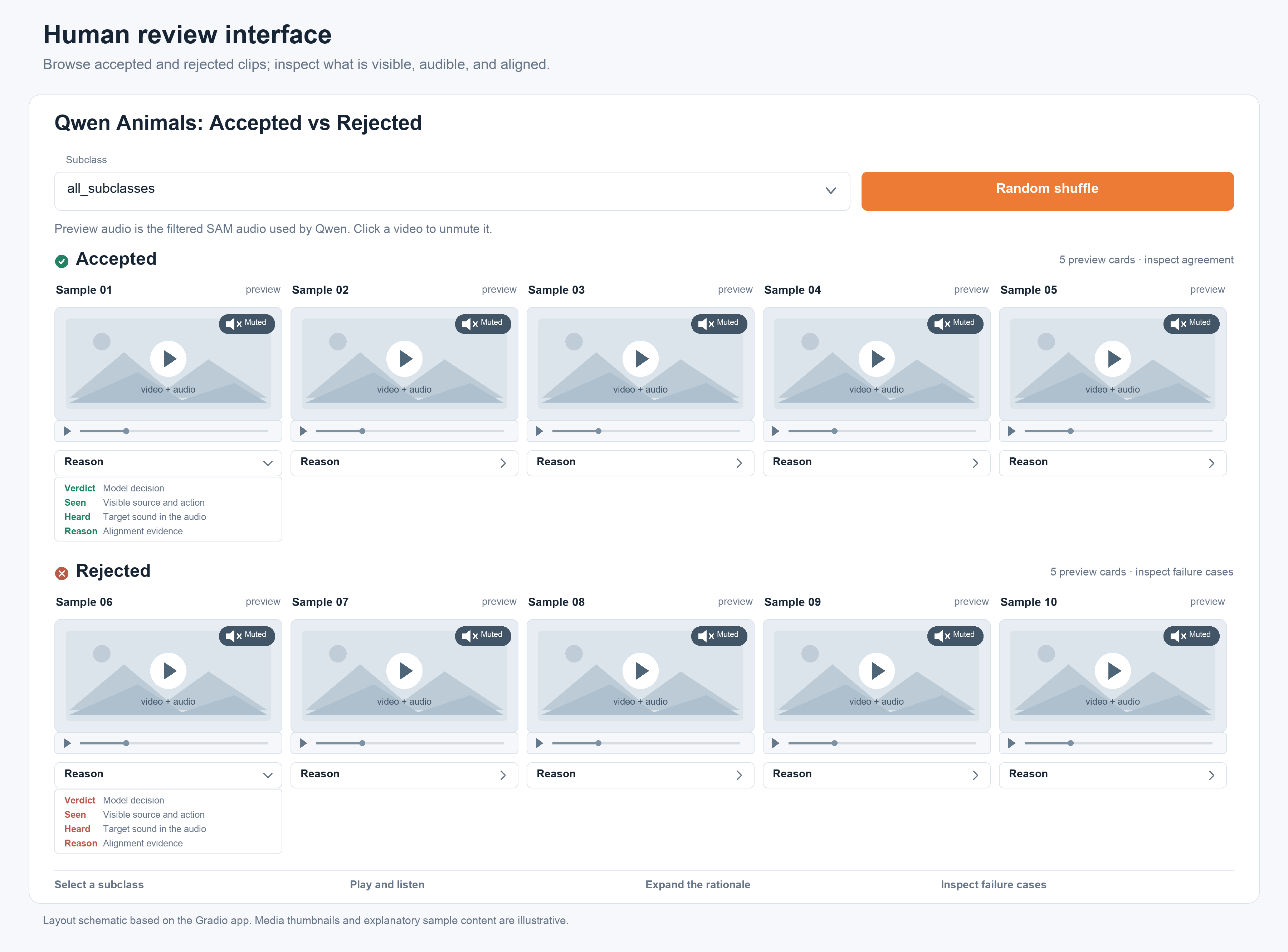}
    \caption{\textbf{Interface for human verification.}}
    \vspace{-1mm}
    \label{fig:human}
\end{figure}

\noindent \textbf{Human-guided filtering}. 
We create gradio interface as shown in Fig.~\ref{fig:human}. And iteratively go over the clips to update our prompt, vocabulary and threshold for different curation stages.

\begin{figure*}[t]
    \centering
    \includegraphics[width=0.95\linewidth]{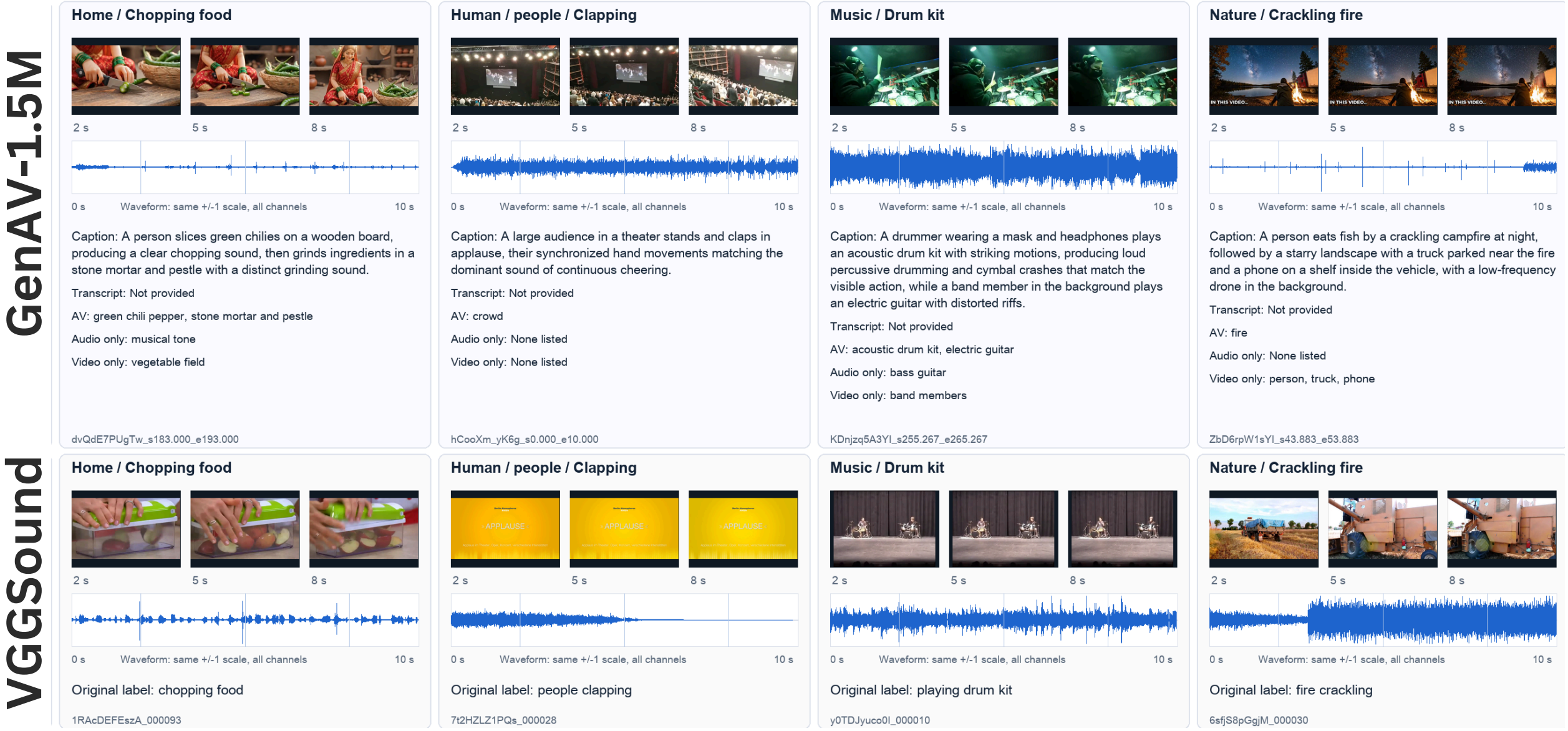}
    \caption{\textbf{Qualitative comparison of {\ourdata} and VGGSound.}
    We show four examples from corresponding sound categories, including video frames and audio waveforms. Unlike VGGSound's class-level labels, {\ourdata} provides detailed audio--visual captions, modality-specific annotations, and speech transcriptions when available. The examples illustrate differences in source visibility, audio--visual correspondence, and annotation richness.}
    \vspace{-1mm}
    \label{fig:qual_example}
\end{figure*}

%%%%%%%%%%%%%%%%%%%%%%%%%%%%%%%%%
\begin{figure*}[h]
    \centering
    \includegraphics[width=0.95\linewidth]{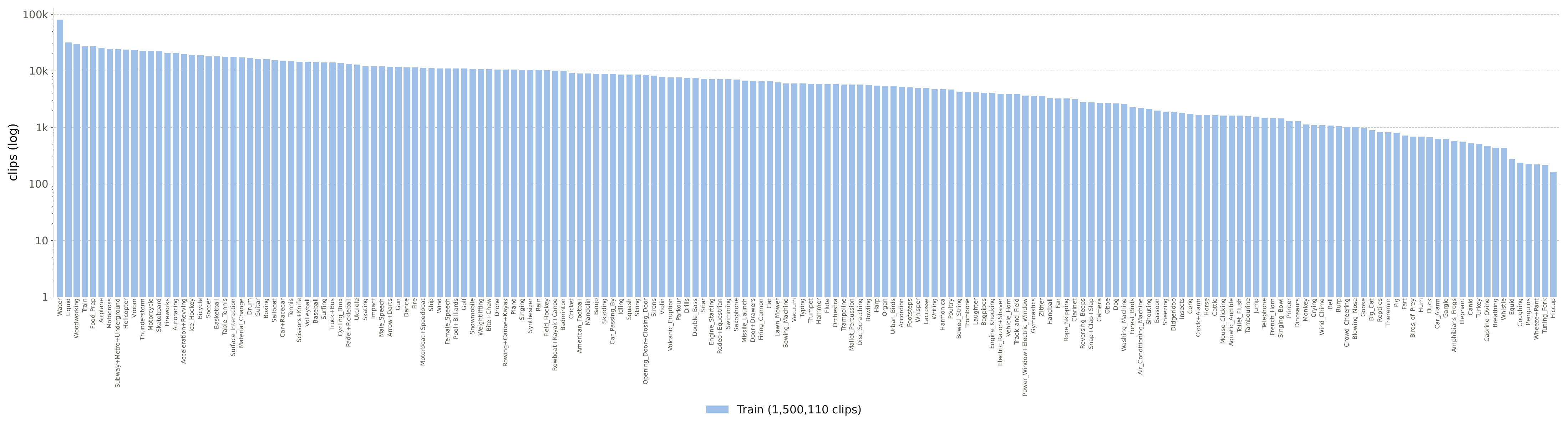}
    \caption{\textbf{Class-wise distribution of the train split of {\ourdata}.}
    Number of clips per sound-source category in the training (1,500,110) shown on a logarithmic scale. Categories are ordered by frequency.}
    \vspace{-1mm}
    \label{fig:class_dist_train}
\end{figure*}
\begin{figure*}[h]
    \centering
\includegraphics[width=0.95\linewidth]{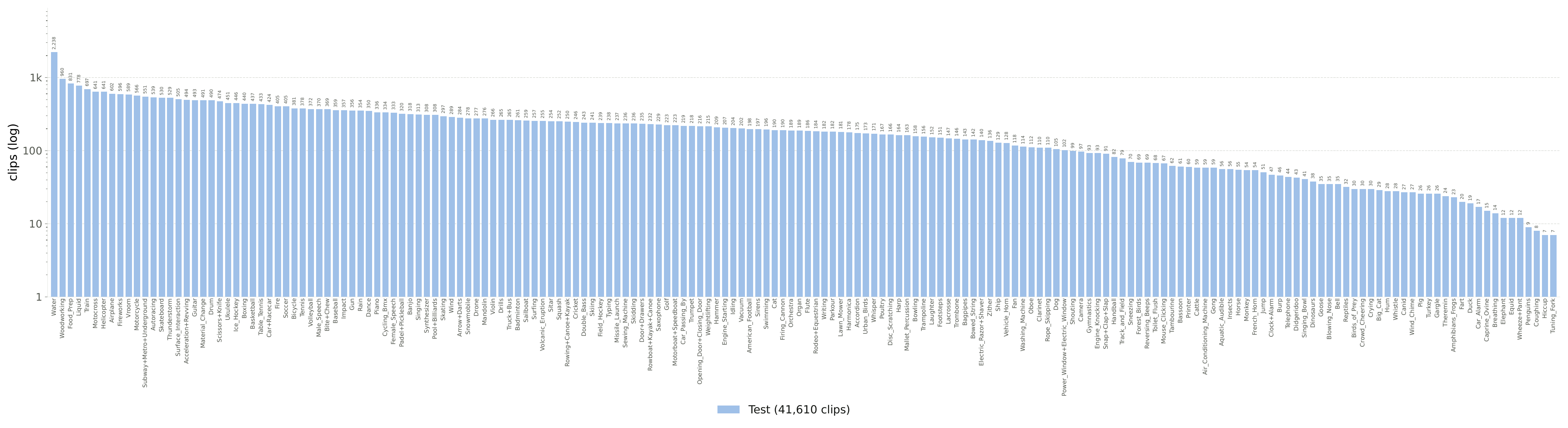}
    \caption{\textbf{Class-wise distribution of the test split of {\ourdata}.}
    Number of clips per sound-source category in the {\ourtestLarge} (41k) shown on a logarithmic scale. Categories are ordered by frequency.}
    \vspace{-1mm}
    \label{fig:class_dist_test}
\end{figure*}
%%%%%%%%%%%%%%%%%%%%%%%%%%%%%%%%%%%%%%%

\section{More about {\ourdata}}

\subsection{Class-wise Distribution}
Fig.~\ref{fig:class_dist_train} and Fig.~\ref{fig:class_dist_test} illustrates the distribution of {\ourdata} across 182 sound-source categories, showing the training and test splits.

\subsection{Qualitative examples}
We present qualitative examples from {\ourdata} and VGGSound in Fig.~\ref{fig:qual_example}, comparing random clips from four sound categories. Our examples exhibit clear audio--visual correspondence, with visible sound-producing sources and actions. In contrast, some VGGSound examples contain title screens or distant views of sound sources, making the correspondence less explicit. Furthermore, while VGGSound provides class-level labels, {\ourdata} includes detailed captions describing sound events, their visual sources, and associated actions, along with modality-specific annotations. These examples highlight the importance of both audio--visual alignment and descriptive annotations for generation-oriented datasets.

\section{Implementation Details}
We build upon recent open-source models for all three tasks.

\noindent \underline{\textit{Video to audio generation}}. We train MMAudio's \texttt{large\_44k} model from scratch using its official training pipeline.

\noindent \underline{\textit{Audio to video generation}}.
For \textit{A2V}, we train an audio-conditioned ControlNet on the frozen Wan2.1-T2V-1.3B backbone. The ControlNet incorporates CLAP~\cite{elizalde2023clap} semantic features and temporally aligned Synchformer~\cite{iashin2024synchformer} audio features while retaining the text conditioning. We train only the ControlNet parameters using a flow-matching objective.

\noindent \underline{\textit{Joint audio-video generation}}.
For \textit{JAV}, we fine-tune pretrained JavisDiT++~\cite{liu2026javisdit++} using rank-64 LoRA during the supervised fine-tuning (SFT) stage, without the subsequent direct preference optimization (DPO)~\cite{rafailov2023direct}. We use 16 FPS videos with 240p aspect-ratio buckets. 

\section{Experiments on {\ourtestLarge}}
We present additional video-to-audio generation experiments on {\ourtestLarge} in Tabs.~\ref{tab:v2a_test_large} and~\ref{tab:v2a_ablate}. The results show trends consistent with those observed on {\ourtest}, with our model trained on {\ourdata} outperforming existing state-of-the-art video-to-audio generation models.
%%%%%%%%%%%%%%%%%%%%%%%%%%%%%%%%%%%
\begin{table}[t]
\centering
\renewcommand{\arraystretch}{1.1}
\setlength{\tabcolsep}{3pt}

\resizebox{\columnwidth}{!}{
\begin{tabular}{lcc|ccccc}

\toprule
\multirow{2}{*}{\bf Model} &
\multirow{2}{*}{\textbf{\#Params}} &
\multirow{2}{*}{\textbf{\#Train}} &
\multicolumn{5}{|c}{\bf {\ourtestLarge}} \\
\cmidrule(lr){4-8}

& & &
\bf FD$_\downarrow$ &
\bf KL$_\downarrow$ &
\bf AT-Sem.$^\uparrow$ &
\bf AV-Sem.$^\uparrow$ &
\bf AV-Sync$_\downarrow$ \\
\midrule

FoleyCrafter & 1.2B & 304K
& 21.09 & 2.06 & 0.26 & 0.29 & 1.21 \\

Frieren & 0.1B & 182K
& 12.10 & 2.3 & 0.19 & 0.24 & 0.80 \\

MMAudio-L & 1B & 1.13M
& 11.89 & 1.98 & \textbf{0.32} & \underline{0.36} & \underline{0.37} \\

Hunyuan-Foley & 5.1B & \faLock
& 15.94 & \underline{1.73} & 0.26 & \underline{0.36} & 0.44 \\

Omni2Sound & 1.1B & 3M
& \underline{8.35} & 1.78 & 0.27 & \underline{0.36} & 0.43 \\

\midrule
\rowcolor{cyan!13}
\textbf{MMAudio-L} (\textit{Ours}) & 1B & 1.5M
& \textbf{5.35} & \textbf{1.41} & \underline{0.29}
& \textbf{0.37} & \textbf{0.34} \\

\bottomrule
\end{tabular}
}

\caption{\textbf{Video-to-audio generation results on {\ourtestLarge}}
We compare existing V2A models with MMAudio-L trained on
\ourdata{}. Our model achieves superior distributional
quality and temporal synchronization.
\faLock{} indicates proprietary training data.}
\label{tab:v2a_test_large}
\end{table}

%%%%%%%%%%%%%%%%%%%%%%%%%%%%%%%%%%%
%%%%%%%%%%%%%%%%%%%%%%%%%%
\begin{table}[t]
\centering
\renewcommand{\arraystretch}{0.95}
\setlength{\tabcolsep}{4pt}

\resizebox{\linewidth}{!}{
\begin{tabular}{l|ccccc}
\toprule
% \rowcolor{gray!15}
% \multicolumn{6}{c}{\textit{Video-to-Audio Generation}} \\
% \midrule

\multirow{2}{*}{\textbf{Train data}} &
\multicolumn{5}{c}{\textbf{{\ourtestLarge}}} \\
\cmidrule(lr){2-6}

& \bf FD$_\downarrow$
& \bf KL$_\downarrow$
& \bf AV-Sem.$^\uparrow$
& \bf AT-Sem.$^\uparrow$
& \bf AV-Sync$_\downarrow$ \\
\midrule

VGGSound
& 9.63 & 1.90 & 0.30 & 0.24 & 0.39 \\
AudioSet-SA
& 11.23 & 1.87 & 0.31 & \textbf{0.26} & \textbf{0.36} \\
\rowcolor{cyan!13}
GenAV-1.5M
& \textbf{6.13} & \textbf{1.41} & \textbf{0.32} & 0.25 & 0.37 \\

\bottomrule
\end{tabular}
}
\caption{
\textbf{Effect of training data across three audio-visual generation tasks.}
We compare models trained on VGGSound, AudioSet-SA,
and GenAV-1.5M, evaluated on {\ourtestLarge}.
All comparisons use task-specific evaluation metrics.
}
\label{tab:v2a_ablate}

\end{table}
%%%%%%%%%%%%%%%%%%%%%%%%%%

\section*{Limitations and Future Work}
Following prior datasets~\cite{chen2020vggsound,liu2025omniaudiogeneratingspatialaudio}, we plan to release YouTube URLs and timestamps rather than redistributing videos. As the content rights belong to the original uploaders, some videos may become inaccessible over time. Like other large-scale web-sourced datasets, {\ourdata{}} may inherit source-platform biases and memorization of training data. Despite rigorous filtering, our curation pipeline remains dependent on the capabilities of the underlying models, and could benefit from future model improvements. 

While this work focuses on audio-visual generation, the dataset could also support tasks such as audio-visual question answering, captioning, and sound source localization. Furthermore, our scalable pipeline can be extended to additional classes and larger datasets with minimal human verification.

\section*{License}
The videos used in this study were sourced from publicly accessible YouTube content. All content is copyrighted by their respective creators and owners. 
We follow YouTube's Terms of Service and applicable licensing requirements, using content permitted for research or under fair-use considerations.

\end{document}